\documentclass{article}

\usepackage{multirow}
\usepackage{multicol}
\usepackage{siunitx}
\usepackage{bm}
\usepackage{wrapfig}
\usepackage{subfig}
\usepackage{listings}
\usepackage{xcolor}
\usepackage{tikz}
\usepackage{tabularx}   
\usepackage{tabularray} 
\usepackage{graphicx}
\usepackage{float}
\usepackage{subcaption} 

\usepackage{authblk}
\lstdefinelanguage{C++}{
  language=C,
  morekeywords={
    class, public, private, protected, template, typename,
    std, cout, cin, endl, string, vector, map, unordered_map,
    auto, constexpr, noexcept, nullptr, using, namespace, new, delete
  },
  sensitive=true,
  morecomment=[l]{//},
  morecomment=[s]{/*}{*/},
  morestring=[b]",
}

\usepackage[final]{neurips_2025}

\usepackage[utf8]{inputenc} 
\usepackage[T1]{fontenc}    
\usepackage{hyperref}       
\usepackage{url}            
\usepackage{booktabs}       
\usepackage{amsfonts}       
\usepackage{nicefrac}       
\usepackage{microtype}      
\usepackage{xcolor}         
\usepackage{natbib}
\usepackage{amsmath}
\usepackage{xspace}
\usepackage{hyperref}

\newcommand{\ie}{i.e.}

\newcommand{\algname}{Tru-POMDP\xspace}
\newcommand{\roomnode}{\textit{ROOM}\xspace}
\newcommand{\areanode}{\textit{AREA}\xspace}
\newcommand{\objnode}{\textit{OBJECT}\xspace}
\newcommand{\robnode}{\textit{ROBOT}\xspace}
\newcommand{\move}{\textit{MOVE}\xspace}
\newcommand{\pick}{\textit{PICK}\xspace}
\newcommand{\place}{\textit{PLACE}\xspace}
\newcommand{\open}{\textit{OPEN}\xspace}

\newcommand{\none}{\textit{NULL}\xspace}

\newcommand{\apick}{\textit{PICK(area, object)}\xspace}
\newcommand{\aplace}{\textit{PLACE(area)}\xspace}
\newcommand{\aopen}{\textit{OPEN(area)}\xspace}

\newcommand{\stategraph}{\ensuremath{\mathcal{G}_s}\xspace}
\newcommand{\particlegraph}{\ensuremath{\mathcal{G}_{s_i}}\xspace}
\newcommand{\obssg}{\ensuremath{\mathcal{G}_z}\xspace}
\newcommand{\stategoal}{\ensuremath{p_s}\xspace}
\newcommand{\nextgraph}{\ensuremath{\mathcal{G}_{s'}}\xspace}
\newcommand{\nextstategoal}{\ensuremath{p_{s'}}\xspace}
\newcommand{\vis}{\ensuremath{\texttt{Visible}}\xspace}

\newcommand{\llmbelief}{\ensuremath{b_{\text{LLM}}}\xspace}
\newcommand{\bfbelief}{\ensuremath{b_{\text{BF}}}\xspace}
\newcommand{\updatedbelief}{\ensuremath{b_{t'}\xspace}}
\newcommand{\objectset}{\ensuremath{\mathcal{O}}\xspace}
\newcommand{\goalplacementset}[1]{\ensuremath{\mathcal{P}_{\text{goal}}(#1)\xspace}} 
\newcommand{\initset}[1]{\ensuremath{\mathcal{P}_{\text{init}}(#1)\xspace}}        

\newcommand{\llmparticle}{\ensuremath{s}\xspace}

\newcommand{\draft}[1]{\textcolor{black}{#1}}

\newcommand{\modify}[1]{\textcolor{black}{#1}}

\title{
\algname: Task Planning Under Uncertainty via Tree of Hypotheses and Open-Ended POMDPs
}

\author{
\textbf{Wenjing Tang}\textsuperscript{1,2\S},
\textbf{Xinyu He}\textsuperscript{2,3},
\textbf{Yongxi Huang}\textsuperscript{1,2\S},
\textbf{Yunxiao Xiao}\textsuperscript{2,4},
\textbf{Cewu Lu}\textsuperscript{1,2},
\textbf{Panpan Cai}\textsuperscript{1,2\dag} \\
\textsuperscript{1}Shanghai Jiao Tong University \quad
\textsuperscript{2}Shanghai Innovation Institute \\
\textsuperscript{3}East China Normal University \\
\textsuperscript{4}Beijing University of Posts and Telecommunications \\
}

\begin{document}

\maketitle

{\renewcommand{\thefootnote}{\S}
 \footnotetext{Wenjing Tang and Yongxi Huang are the visiting students at Shanghai Innovation Institute.}
}
{\renewcommand{\thefootnote}{\dag}%
 \footnotetext{Corresponding author: \texttt{cai\_panpan@sjtu.edu.cn}}
}
\begin{abstract}
\label{abstract}
Task planning under uncertainty is essential for home-service robots operating in the real world. Tasks involve ambiguous human instructions, hidden or unknown object locations, and open-vocabulary object types, leading to significant open-ended uncertainty and a boundlessly large planning space. To address these challenges, we propose \algname, a planner that combines structured belief generation using Large Language Models (LLMs) with principled POMDP planning. \algname introduces a hierarchical \textit{Tree of Hypotheses} (TOH), which systematically queries an LLM to construct high-quality particle beliefs over possible world states and human goals. We further formulate an open-ended POMDP model that enables rigorous Bayesian belief tracking and efficient belief-space planning over these LLM-generated hypotheses. Experiments on complex object rearrangement tasks across diverse kitchen environments show that \algname significantly outperforms state-of-the-art LLM-based and LLM-tree-search hybrid planners, achieving higher success rates with significantly better plans, stronger robustness to ambiguity and occlusion, and greater planning efficiency.\draft{ \footnote{The code and demonstration video are available at: \url{https://tru-pomdp.github.io}}}


\end{abstract}

\section{Introduction}
\label{intro}

Home-service robots are increasingly expected to perform complex tasks in unstructured household settings, such as tidying up, preparing for guests, or assisting with daily routines. Among these, object rearrangement tasks---e.g., ``prepare the kitchen for a party''---require robots to interpret ambiguous, open-ended instructions and manipulate open-vocabulary objects, many of which may be hidden in drawers, cabinets, or containers. The robot must infer user intent, identify relevant items, and plan long-horizon action sequences to satisfy under-specified goals in partially observable environments. 


\modify{While these challenges can be positioned within the framework of planning under uncertainty---where a robot must hedge against imperfect perception, unobservable environment states, and stochastic action outcomes---the difficulties in open household environments extend far beyond standard formulations. First, human instructions are inherently ambiguous and often omit crucial details (e.g., what specific items are needed for a party). Second, many objects are hidden from the robot's initial view, and the set of relevant objects is unbounded and diverse (e.g., cups, snacks, decorations). As a result, the robot must reason over \textit{open-ended uncertainty}, where fundamental components of the planning problem---such as the state space (e.g., what objects exist), the action space (e.g., what objects to operate on) and task goals (e.g., what the human want)---cannot be predefined. This setting leads to an effectively boundless belief space, making long-horizon planning extremely challenging.}

\modify{To address open-world problems}, recent work has started to leverage large language models (LLMs) to bring commonsense reasoning into robot planning. One line of work treats LLMs as planners, leveraging their implicit knowledge and ability to reflect over past failures~\cite{yao2022-react, shinn2023-reflexion, huang2022inner_LLM-Monologue-Planners, huang2022language_LLM-ZeroShot-Planners, ahn2022can, liang2023code}. However, such approaches often rely on implicit, unverified hypotheses about the world, and can fail catastrophically when the underlying hypotheses are incorrect. Another line combines LLMs with symbolic or probabilistic planners, for example, by prompting LLMs to generate POMDP models for informing LLM planning or suggesting default actions for search ~\cite{zhao2023large-LLM_MCTS, Lee-PrimeTheSearch, Sun-LLM_POP, xiong-etal-2024-watchEveryStep, curtis2025llmguidedprobabilisticprograminduction, chen2025robohorizonllmassistedmultiviewworld}. While more structured, these methods still commit to a single high-probability hypothesis or \modify{operate in a closed domain, which limits robustness in the open-world.}

\modify{To overcome these limitations, we propose \algname, a task planner built upon an \textit{open-ended POMDP} formulation, that tightly integrates commonsense reasoning by LLMs with \emph{explicit} belief tracking and \emph{principled} POMDP planning.}
\algname operates in three stages: (1) it constructs a tree of hypotheses over possible world states and task goals using hierarchical LLM queries, to form a commonsense belief; (2) it fuses the LLM-generated belief with Bayesian filtering to construct a hybrid belief that is both diverse and consistent during updates; and (3) it solves for an optimal policy using online belief tree search with a dynamically constructed action space and guided by an LLM-generated rollout policy. This design leverages the generalization power of LLMs while maintaining rigorous, verifiable reasoning over open-ended uncertainty.

We evaluate \algname on object rearrangement tasks across five diverse kitchen environments from RoboCasa~\cite{nasiriany2024robocasa}, involving open-ended instructions, ambiguous user intentions, and a large number of hidden spaces. 
Results show that \algname effectively tackles challenges brought by open-ended uncertainties. Through the integration of LLM-based belief generation with POMDP planning, \algname significantly outperforms both LLM agents and LLM-augmented tree search planners, producing higher success rates and plan qualities with lower LLM token usage. \algname also benefits from structured belief modeling with LLMs and hybrid belief tracking that combines LLM belief with Bayesian updates.

In summary, our main contributions include:
\begin{itemize}
\item The first framework to address ``open-ended POMDPs'', integrating LLM-based reasoning with \textit{principled} POMDP planning for household tasks.
\item A novel hybrid belief modeling approach that integrates LLM-generated hypotheses with \textit{principled} Bayesian filtering.
\item A POMDP model for {open-world} object rearrangement tasks and a practical belief-tree search planner for solving such tasks efficiently under large-scale uncertainty.
\end{itemize}

\section{Background and Related Work}
\label{gen_inst}

\subsection{Online POMDP Planning}

Partially Observable Markov Decision Processes (POMDPs) provide a principled framework for decision-making under uncertainty, where the true state of the world is only partially and noisily observed. A POMDP is formally defined as a 7-tuple \((S, A, Z, T, O, R, b_0)\), where \(S\) is the set of states, \(A\) is the set of actions, \(Z\) is the set of observations, \(T(s'|s,a)\) is the transition model, \(O(z|s',a)\) is the observation model, \(R(s,a)\) is the reward function, and \(b_0\) is the initial belief over states.
Online POMDP planning targets for single-query planning or even real-time planning, by computing policies via \textit{belief tree search}. The agent maintains a belief---a probability distribution over possible world states---using Bayesian filtering. At each decision step, it constructs a belief tree by simulating future actions and observations \cite{NanYe2017DESPOT}, and applies Bellman's principle \cite{kaelbling1998planning} in the tree to compute the best policy given the current belief. Only the immediate action is executed, and the process is repeated at each time step, allowing the planner to adapt to new observations in an online fashion (e.g., replan at 1 Hz). \modify{A POMDP model requires a full specification of all possible states, actions, and observations, which limits its use to "closed-domain" problems.}

\subsection{Robot Planning with Large Language Models}

Recent advances in large language models (LLMs) have enabled robots to handle tasks expressed in open-vocabulary language and to incorporate commonsense knowledge into the planning process. Broadly, two directions have emerged. The first uses LLMs \textit{as planners}, either generating open-loop action sequences via prompt engineering or fine-tuning~\cite{huang2022language_LLM-ZeroShot-Planners, huang2022inner_LLM-Monologue-Planners, singh2023progprompt, chen2024agent, ahn2022can, liang2023code, zhou2024isr}, or producing closed-loop actions through iterative feedback and reflection~\cite{shinn2023-reflexion, yao2022-react, Sun-AdaPlanner, wang2024learning-NAT, song2024-ETO, yang2025selfgoal, raman2022planning, bhat2024grounding}. While flexible, these approaches rely on implicit reasoning and often lack robustness when their internal hypotheses diverge from reality. The second direction combines LLMs with \textit{explicit planning algorithms}. For example, some work uses LLMs to generate PDDL domains and problems, which are then solved by symbolic planners~\cite{liu2023-LLMP, han2024-interpret, Birr2024-AUTOGPT_P, dagan2023dynamic, guan2023leveraging}. Others combine LLMs with Monte Carlo Tree Search (MCTS), using the LLM to hypothesize initial states or initial solutions~\cite{zhao2023large-LLM_MCTS, Lee-PrimeTheSearch, qiao2024-agent, shi2025MC_DML}. However, these methods generally assume full observability and deterministic goals, ignoring the inherent uncertainty in both human intention and world states. As a result, their solutions are often suboptimal in real-world deployments.

\subsection{Planning under Uncertainty with Large Language Models}

A few recent efforts have explored combining LLM reasoning with the structure of POMDPs. Some approaches use LLMs to generate elements of a POMDP model---such as beliefs or observation likelihoods---which are then fed back into the LLM to improve its planning ability under uncertainty~\cite{Sun-LLM_POP, xiong-etal-2024-watchEveryStep, curtis2025llmguidedprobabilisticprograminduction, chen2025robohorizonllmassistedmultiviewworld}. However, these methods still rely heavily on the LLM's implicit reasoning and lack the formal guarantees provided by explicit belief tracking and tree-search planning. 
\modify{Other work combines LLMs with tree search}. One possibility is to generate a single most likely hypothesis about goals and world states, and then applies MCTS for planning~\cite{hazra2024saycanpay, ren2023robotsaskhelpuncertainty}. This strategy fails when the true goal diverges from the most likely one. 
\modify{LLM-MCTS~\cite{zhao2023large-LLM_MCTS} proposed to plan with LLM-generated state distributions. However, it uses a "flat" belief that only considers a single aspect of uncertainty on the initial placement of objects. It operates in a closed domain, assuming the object set is known beforehand, and incurs significant computational cost due to repeated LLM queries within the tree search~\cite{he2024wordsactionsunveilingtheoretical}.
In contrast, \algname offers a scalable and flexible approach to address open-ended uncertainty.} It uses the LLM to construct a \textit{tree of plausible hypotheses} representing a complex, multi-aspect, and open-ended belief over entirely unknown objects and ambiguous human goals. We combine this with Bayesian filtering to maintain a consistently updatable belief. Finally, we apply belief tree search over this hybrid belief, using a {dynamically constructed action space} and a {pre-compiled, LLM-generated rollout policy} to efficiently guide optimal search. 

\section{Problem Formulation}
\label{problem_formulation}

\subsection{Task Specification}

We study the object rearrangement task performed by a dual-armed mobile robot in a household environment. Given a free-form, ambiguous natural language instruction \( I \), the robot must arrange open-vocabulary objects existing in the scene into a target configuration that fulfills the human's underlying intention, while minimizing total execution time. The robot executes four types of actions: \move navigates to a specific area (e.g., in front of furniture or appliances); \pick picks an object from an open surface or an open container; \place places the held object onto a surface or into a container; and \open opens a container or drawer if it is closed. \move has a time cost proportional to the navigation distance, while the other actions incur a fixed cost. The robot can hold at most one object in its right hand; the left hand is reserved for
the \open operation.

The environment is represented as a structured scene graph \(\mathcal{G} = (V, E)\), where \(V\) denotes asset nodes and \(E\) encodes spatial relationships. Nodes are organized hierarchically into four layers: a root \roomnode node representing the overall scene; \areanode nodes representing surfaces or containers that can hold objects; \objnode nodes representing physical items; and a \robnode node indicating the robot's current location. Edges encode containment and adjacency (e.g., an object is at an area, or the robot is at an area). Each \areanode node has a boolean attribute indicating whether it is open or closed. The robot can observe objects in open areas without noise but cannot observe contents in closed areas. While we assume noise-free observations, the method naturally extends to stochastic cases without affecting belief tracking or planning algorithm design.

Due to the inherent ambiguity of human instructions, we model the task goal as a set of plausible placement goals \(\mathcal{P} = \{p_1, \ldots, p_n\}\). Each placement goal \(p_i\) specifies a list of target objects \(\{o_1, \ldots, o_m\}\) and their respective desired destination areas \(\{t_{o_1}, \ldots, t_{o_m}\}\). These specifications are concise and do not constrain the locations of unrelated objects. Concrete goal representations for example tasks are provided in the appendix~\ref{task generation}.

\subsection{The POMDP Model} \label{pomdp_model}

\textbf{States.} A state \(s \in S\) consists of a scene graph \stategraph encoding the current placement of both visible and hidden objects, and a hypothetical placement goal \stategoal of target objects, representing the intended final configuration consistent with the instruction of the task.

\textbf{Observations.} An observation \(z \in Z\) is a partial scene graph \obssg that includes only visible objects in open areas. The observation function is defined as:
\begin{equation}
O(s', a, z) = \Pr(\obssg \mid s' = (\nextgraph, \nextstategoal)) = 
\begin{cases} 
1, & \text{if } \obssg = \vis(\nextgraph) \\ 
0, & \text{otherwise}
\end{cases}
\label{observation-function}
\end{equation}
where \vis{} removes all hidden objects from \nextgraph{}.

\textbf{Parameterized Actions and State Transition.} The action space \(A\) is defined over parameterized operations on the scene graph. \aopen opens a closed area, setting its status to open. \apick picks an object from an area, reassigning its parent to the \robnode node. \aplace places the held object into the target area, making that area its new parent. \draft{These actions implicitly execute a \move when the robot is not in front of the specified area or object.} The robot may also choose to take no action via \none. Actions are subject to feasibility constraints: for instance, only closed areas can be opened. Infeasible actions are mapped to \none.

\textbf{Reward Modeling.} Manipulation actions incur a fixed cost of 5, and navigation costs vary from 0 to 27 depending on the distance traveled. Infeasible actions are penalized with a cost of 100. Completing a subgoal yields a reward of 200, with an additional 200 granted upon full task completion.

\textbf{Belief Modeling.} The belief \(b\) is represented as a weighted particle set \(\{(s_i, w_i) \mid s_i \in S, w_i \in (0, 1]\}\), where each \(s_i\) is a possible world state and \(w_i\) its associated probability. The initial belief is generated from the natural language instruction \(I\) and the initial observation \obssg using LLMs, and updated using Bayesian filtering. Details of belief tracking are provided in Section~\ref{sec:tree_of_hypothesis}.

\section{The \algname Planner}
\label{method}

\begin{figure}
    \centering
    \includegraphics[width=1\linewidth]{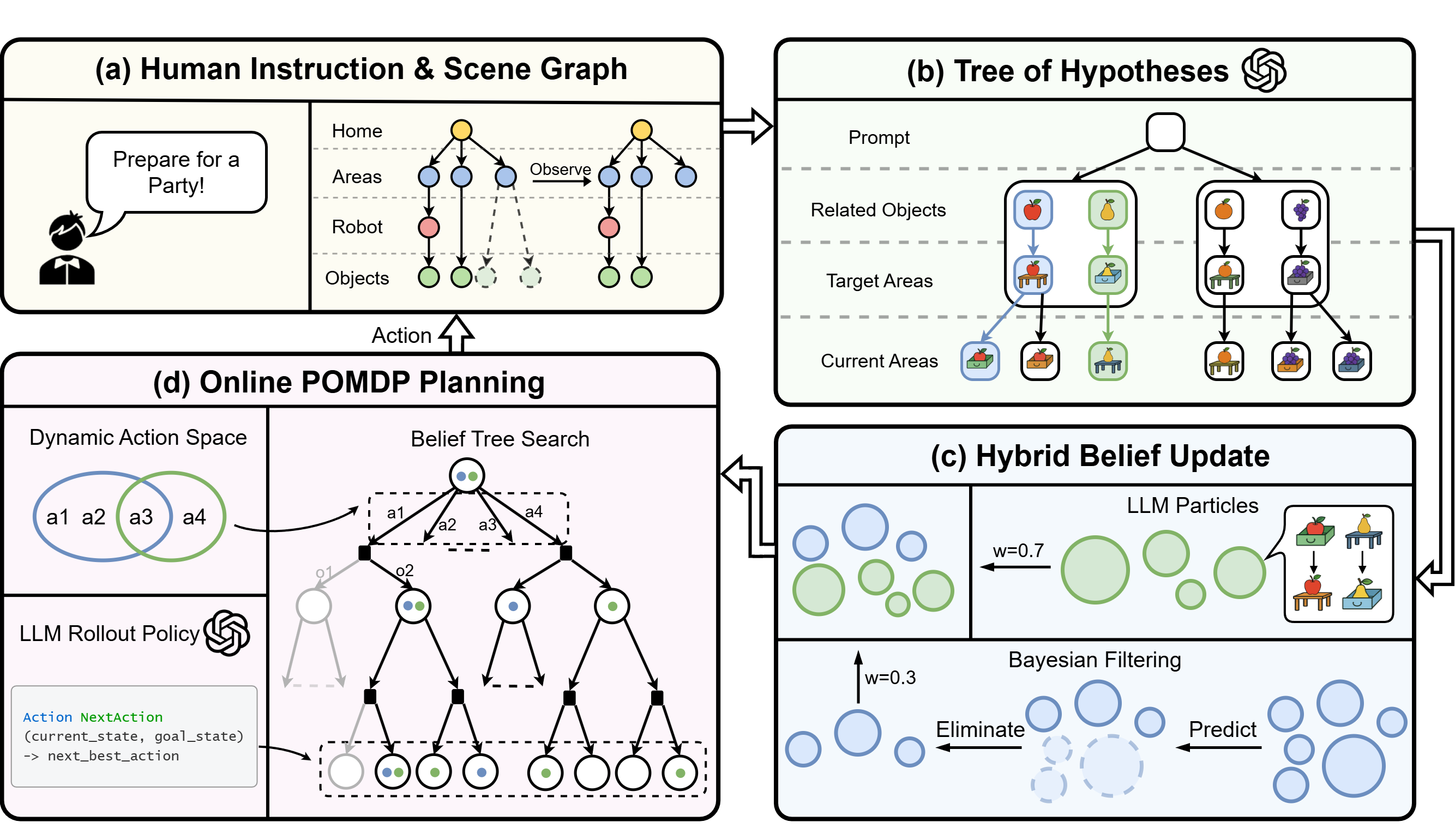}
    \caption{The architecture for \algname. \textit{(a) Task Input}: Human instruction and the observed scene graph. \textit{(b) Tree of Hypotheses}: An LLM infers target objects, target areas, and initial locations, producing weighted particles. \textit{(c) Hybrid Belief Update}: Bayesian filtering updates the belief using particle prediction and elimination, and augments the filtered belief with LLM particles. \textit{(d) Online POMDP Planning}: Belief tree search computes the optimal action with the help of dynamic action branching and an LLM-written rollout policy.}
    \label{fig:Pinepine}
    \vspace*{-0.5cm}
\end{figure}

\algname consists of three modules (Fig.~\ref{fig:Pinepine}): Tree of Hypotheses, Hybrid Belief Update, and Online POMDP Planning. Given a natural language instruction \(I\) and current observation \obssg, the Tree of Hypotheses module generates a belief over candidate world states and plausible placement goals \(\mathcal{P}\). The Hybrid Belief Update module filters inconsistent hypotheses and augments the belief as needed. The Online POMDP module computes the optimal next action under uncertainty via belief tree search.

\subsection{Tree of Hypotheses}
\label{sec:tree_of_hypothesis}

This module queries an LLM to generate a diverse belief over three sources of uncertainty: ambiguity in human instructions, open-class objects, and occluded object locations. The LLM takes as input the human instructions $I$ and the observed scene graph \obssg, transformed into a textual description \(T_z\), and constructs a tree of hypotheses (TOH) consisting of three levels:

\textbf{Level 1: Hypotheses of target objects.}
The LLM predicts sets of objects that are potentially relevant to the task described by \(I\).
\draft{Since we do not assume a known object set, the LLM has to hypothesize objects with commonsense priors, }
and returns a set of \(C_1\) candidate combinations of target objects:
\begin{equation}
\objectset = \{(O_i, w_i) \mid i = 1, \ldots, C_1\},
\end{equation}
where each \(O_i\) is a set of objects that can be used to accomplish the task, and \(w_i\) is the LLM's confidence score for that hypothesis.
These hypotheses reflect alternative interpretations of vague or underspecified instructions. The LLM also incorporates information from failed past executions (if available) to disambiguate user intent and generalize to open-class object names.

\textbf{Level 2: Hypotheses of placement goals.}
For each combination of target objects, the LLM infers a destination area for each object in the set. This yields a plausible goal placement:
\begin{equation}
\goalplacementset{O_i} = \{(o_j, t_j) \mid j = 1, \ldots, |O|\},
\end{equation}
which assigns each target object $o_j$ in the combination $O_i$ to its target \areanode node $t_j$.

\textbf{Level 3: Hypotheses of current placements.}
To estimate the current locations of target objects that are not visible in \obssg, an LLM is queried for each invisible target object \(o\) to predict a distribution over its possible locations, \ie, hidden areas:
\begin{equation}
\initset{o} = \{(l_{o}, w_k) \mid k = 1, \ldots, C_2\}.
\end{equation}
where a set of $C_2$ locations ($l_{o}$'s) are generated, together with a confidence score $w_k$ for each hypothesis.
If \(o\) is observed in \obssg, its location is taken directly. 
 
Inference for each invisible target object is queried independently, enabling parallelization.

\textbf{Particle belief construction.}
Each path from the root to a leaf of the hypothesis tree defines a sampled particle \(s = (\stategraph, \stategoal)\), where \stategraph{} encodes a hypothesized complete scene graph (a union of observed scene graph \obssg and inferred initial placements of invisible target objects), and \stategoal{} corresponds to the generated placement goal. Aggregating all such particles yields the LLM-inferred particle belief:
\begin{equation}
\llmbelief = \{(\llmparticle_n, w_n) \mid n = 1, \ldots, C_3\}.
\end{equation}
where the particle weight $w_n$ is proportional to the multiplication of all confidence scores, \ie, $w_i$'s and $w_k$'s along the top-down path.
These particle weights are normalized to ensure the belief forms a valid probability distribution.

\subsection{Hybrid Belief Update}

The Hybrid Belief Update module maintains a consistent and efficient particle belief over partially observed world states and placement goals. It fulfills two objectives: (1) ensuring smooth and reliable belief updates through Bayesian filtering, and (2) reducing the frequency of expensive LLM queries by only invoking them when necessary. The algorithm monitors the total weight of the particle belief. If the belief retains sufficient weight---indicating strong agreement with the current observation---it is updated using a particle filter. If the weight falls below a threshold (e.g., \(1 - \epsilon = 0.3\)), new particles are generated by invoking the Tree of Hypotheses module (Section~\ref{sec:tree_of_hypothesis}) to replenish the belief.

\textbf{Bayesian Filtering.}
We employ a particle filter to perform principled belief updates. Each step consists of two phases:

\textit{Prediction step.} For each particle \(s_i = (\particlegraph, p_{s_i})\), we simulate the effect of the last action to update the scene graph \particlegraph according to the deterministic transition model.

\textit{Elimination step.} Given the new observation \obssg, particles inconsistent with it are removed. Because the observation model is deterministic (Equation~\ref{observation-function}), this reduces to filtering out particles whose predicted visible scene \(\vis(\particlegraph)\) does not match \obssg. The result is a new filtered particle set \bfbelief with total weight \(w_{\text{BF}} \leq 1\).

\textbf{LLM Particle Supplementation.}
When \(w_{\text{BF}} < 1 - \epsilon\), the belief is augmented using a new set of particles generated by the Tree of Hypotheses. This new set, \llmbelief, is generated by conditioning on the full history of past actions and observations. The augmented belief is defined as:
\begin{equation}
\updatedbelief = \bfbelief + (1 - w_{\text{BF}}) \cdot \llmbelief,
\end{equation}
where the weights of particles in \llmbelief are scaled by \((1 - w_{\text{BF}})\) before merging. This ensures that the resulting belief remains a valid probability distribution.

\subsection{Online POMDP Planning}
\label{sec:Online_POMDP_Planning}

This module plans the optimal policy given the current belief over uncertain world states and goals. Our planner is built upon DESPOT~\cite{NanYe2017DESPOT}, a sampling-based online POMDP algorithm that provides asymptotic optimality guarantees. We extend DESPOT with two key innovations: (1) a dynamic action space constructed from belief particles, and (2) an LLM-generated rollout policy that injects commonsense domain knowledge while maintaining computational efficiency. \algname operates in an online planning setting: at each time step, it executes the optimal action for the current belief and replans after receiving new observations. 

\textbf{Dynamic Action Space.}
As discussed in Section~\ref{pomdp_model}, the robot's action space includes \pick actions over objects that may be hidden and unknown. Since the full set of possible objects is unbounded, naively enumerating all possible parameterized actions would result in an intractably large—or infinite—action space. To address this, we dynamically construct a compact action space based on the current belief. Concretely, we extract the union of all relevant entities from the particles in the belief: hypothesized target objects, open areas, and closed containers. We then construct the grounded action set consisting of: (1) \open for each closed area; (2) \pick for each target object that is currently visible and graspable; and (3) \place for placing the held object (if any) into each open area. This dynamic action set varies with the belief but remains compact and tractable, while retaining sufficient expressiveness for solving the task.

\textbf{Belief Tree Search.}
We adapt the DESPOT algorithm to recursively explore the space of future action-observation sequences from the current belief. At each time step, a sparse belief tree is constructed: nodes represent future belief states (particle beliefs), and branches represent possible actions and the resulting observations, simulated using the POMDP transition and observation models.
Each belief node also maintains a value estimate, representing the expected cumulative reward that can be achieved under optimal behavior conditioned on that belief. These values are updated iteratively using Bellman's principle~\cite{NanYe2017DESPOT}. The planner leverages value estimates as tree search heuristics: it selects actions that maximize the value estimate, and expands observation branches based on excess uncertainty. This ensures targeted exploration of promising belief trajectories, ensuring efficient convergence of values and allowing planning to terminate at any time and output high-quality decisions.

\textbf{LLM-Generated Rollout Policy.}
Following the DESPOT framework, leaf nodes in our belief tree require rolling-out a default policy to provide a heuristic lower bound on their value. However,
unlike prior work that simply uses a random rollout policy~\cite{zhao2023large-LLM_MCTS},
we query the LLM to synthesize a rollout policy in code.
\modify{The rollout policy is a mapping from the current state $s=(G_s,p_s)$ to an action $a$, where $G_s$ is the scene  graph representing the hypothesized world state, and $p_s$ is the placement goal. This mapping is implemented as a C++ function generated by an LLM. The generator LLM receives a general language description of the object rearrangement domain but is given no information about any specific task instance.}
The resulting policy generalizes across tasks in the same domain and is reused throughout the search. During rollouts, this policy is simulated for each rolled-out node to estimate cumulative rewards from the leaves. 
\modify{See Appendix~\ref{rollout policy prompt} for the generator prompt and the code of the rollout policy.}

\section{Experiments} 
\label{experiments}

We conduct comprehensive experiments to evaluate the effectiveness of \algname in open-ended object rearrangement tasks under uncertainty. Our study aims to answer the following questions:
\begin{enumerate}
    \item Is belief-space planning essential for household tasks involving partially observable environments and ambiguous free-form instructions?
    \item Can \algname effectively solve household tasks with high and open-ended uncertainty?
    \item What are the benefits of integrating belief-space planning with LLM-based reasoning?
    \item How does combining Bayesian filtering with LLM belief generation improve performance?
    \item Does the proposed Tree of Hypotheses (TOH) structure generate high-quality beliefs?
\end{enumerate}

\subsection{Experimental Setup}
\label{experimental_setup}

We evaluate \algname in five kitchen environments from RoboCasa~\cite{nasiriany2024robocasa}\draft{, which highlights a large number of hidden areas}: \textit{One Wall}, \textit{One Wall with Island}, \textit{L-Shaped}, \textit{L-Shaped with Island}, and \textit{Galley}. Each scene contains up to 40 \areanode nodes \draft{with semantically-rich names}, of which up to 29 are initially closed.
Tasks are procedurally generated using an LLM-assisted pipeline\draft{, emphasizing vague task instructions with substantial implicit information and ambiguous object references}. For each task, we populate up to 20 objects (including distractors) in the scene using commonsense priors, and define a goal set \(\mathcal{P}\) containing \draft{various} plausible target object combinations and goal placements grounded in the scene. This setup ensures each task is both solvable and exhibits ambiguity in human intention. Example scene with instruction and goal sets are listed in Appendix \ref{kitchen env appendix} and \ref{task generation}.

We categorize tasks into three difficulty levels based on the number of target objects required in the goal: \textit{easy} (requiring 2 target objects), \textit{medium} (3), and \textit{hard} (4–8). 
\draft{Each additional target object in the goal leads to exponential growth of uncertainty.}
For each level, we generate 100 tasks, resulting in a total of 300 tasks.
Step limits are set to 25, 30, and 35 respectively. The planning time is capped at 600 seconds. A task fails if either the step or time limit is exceeded.

We evaluate performance using the following metrics. (1) \textit{Cumulative reward:} the total reward collected in an episode, computed using the POMDP reward function described in Section~\ref{pomdp_model}. In experiment, we remove the reward for subgoals, and adapt the reward for full task completion to 1,000. In addition, we adapt the cost of infeasible actions to 25.
(2) \textit{Success rate:} the percentage of episodes in which the task is successfully completed within the step and time limits. (3) \textit{Step count:} the number of execution steps per episode, where failed episodes are assigned the maximum allowed steps. (4) \textit{Planning time:} the total wall-clock time spent in online planning across all steps of an episode. (5) \textit{LLM token usage:} the total number of input and output tokens consumed by the planner, averaged per episode.


Experiments are run on a local machine equipped with a 12th Gen Intel\textsuperscript{\textregistered} Core\texttrademark~i7-12700KF CPU (20 threads), without GPU acceleration. All methods consistently use GPT‑4.1 as the LLM. 

\subsection{Comparison Results}
\label{comparison_results}

We compare \algname against a set of strong baselines, including both pure LLM-based planners and tree search planners integrated with LLM reasoning. 
\begin{figure}[t!]
    \centering
    \includegraphics[width=0.9\textwidth]{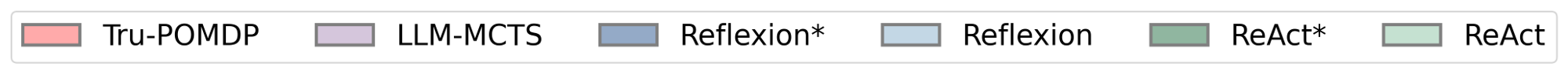}\\[0.5em]
    {\footnotesize 
    \begin{tabular}{@{\hskip 2pt}c@{\hskip 2pt}c@{\hskip 2pt}c@{\hskip 2pt}}
        \includegraphics[width=0.33\textwidth]{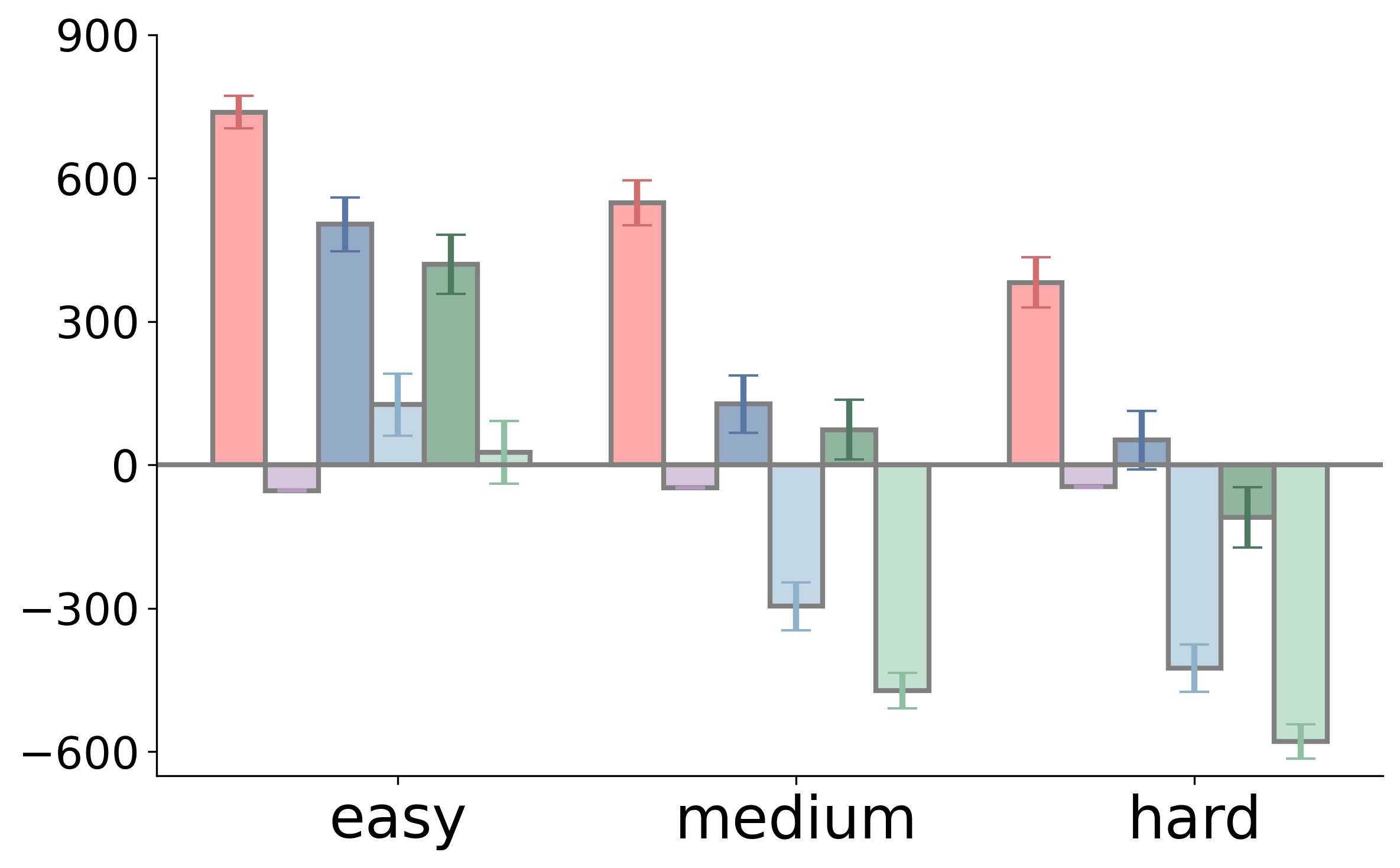} &
        \includegraphics[width=0.33\textwidth]{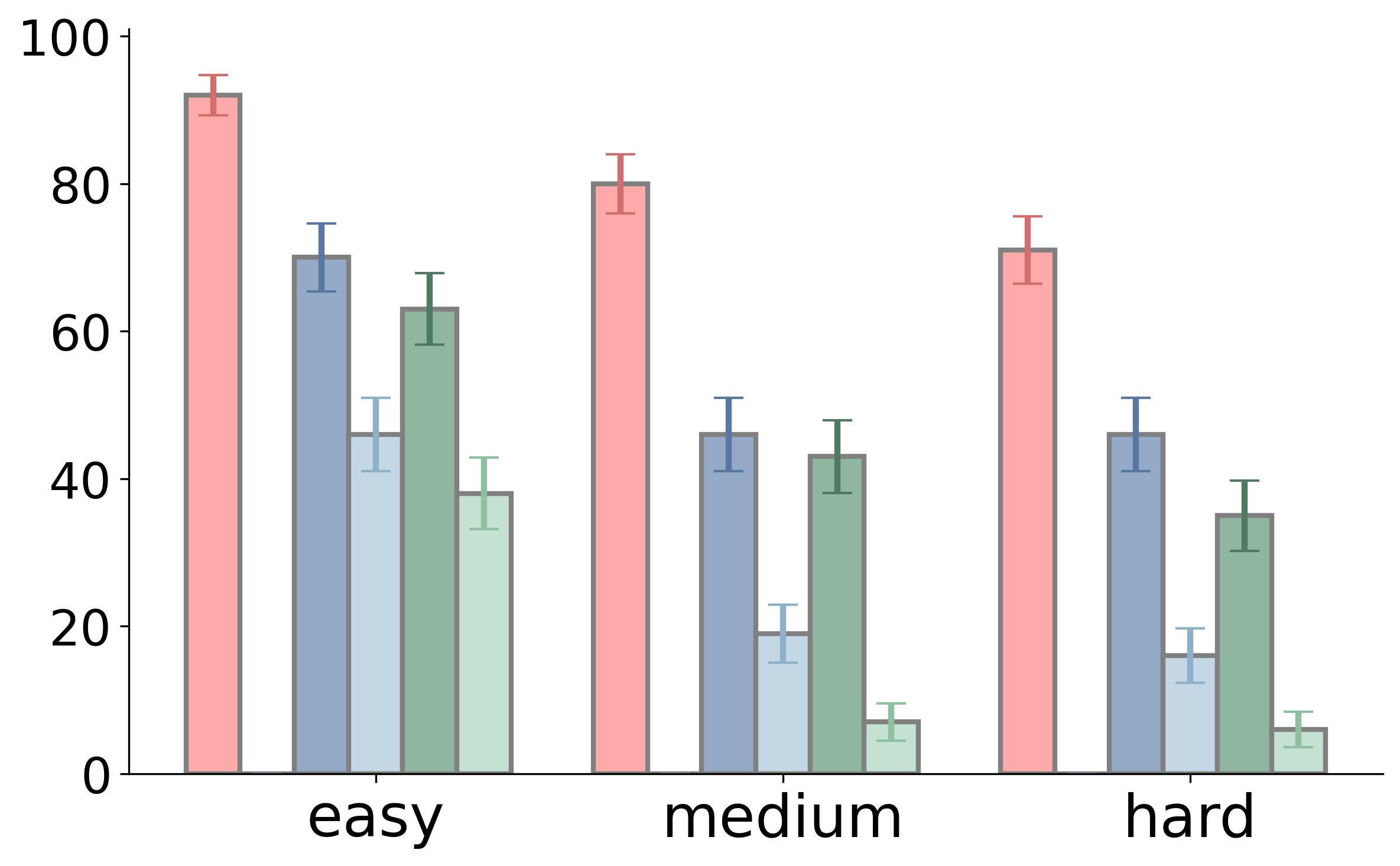} &
        \includegraphics[width=0.33\textwidth]{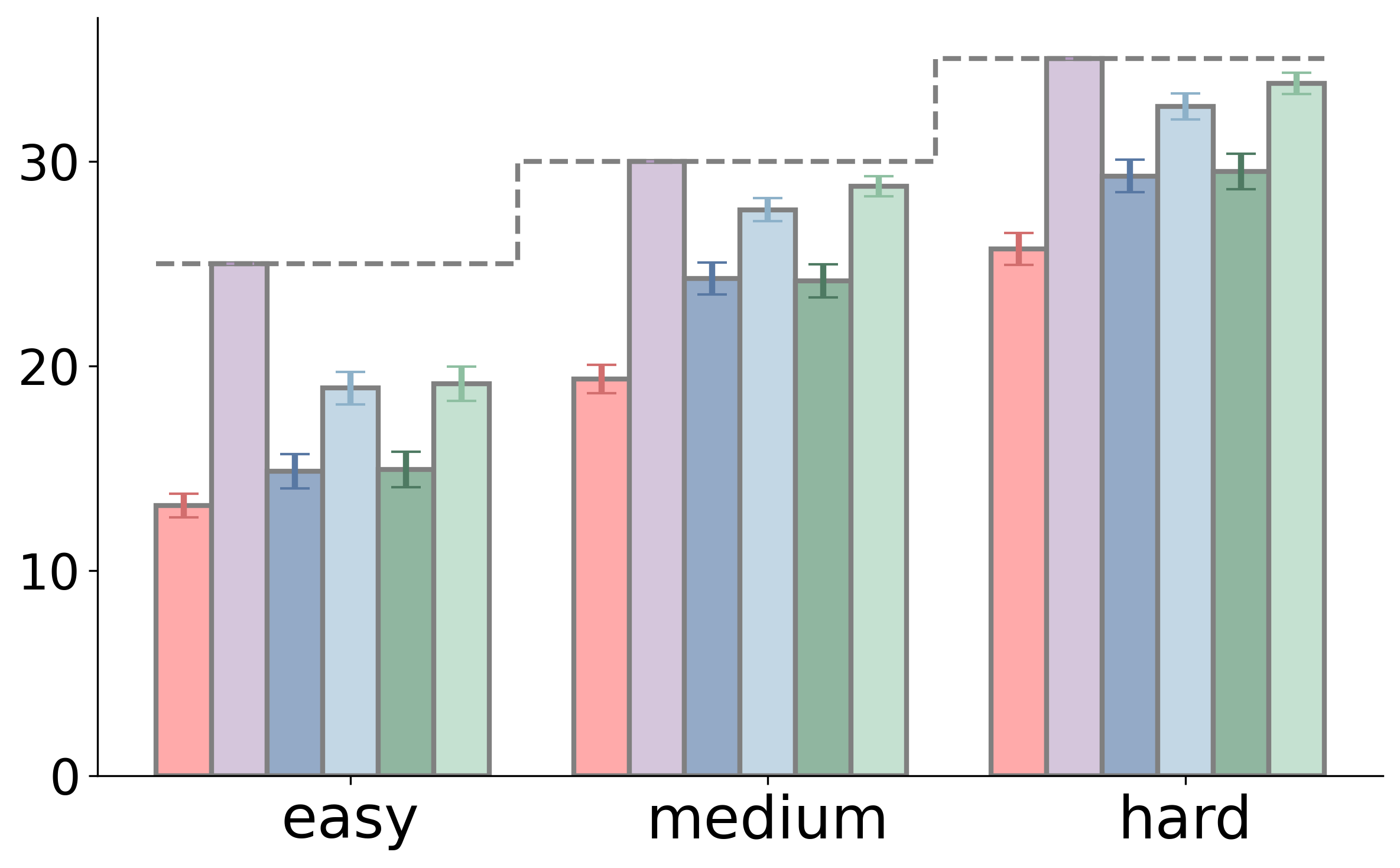} \\
        (a) Cumulative Reward $\uparrow$ &
        (b) Success Rate (\%) $\uparrow$ &
        (c) Total Step Number $\downarrow$
    \end{tabular}
    }
    \caption{Performance comparison of \algname and baselines. Each bar represents the average value with standard error (SE). In (c), the dashed line indicates the maximum allowed step number.}
    \label{total results}
        \vspace*{-0.5cm}

\end{figure}
\begin{itemize}
    \item \textit{ReAct}~\cite{yao2022-react}: A closed-loop LLM-based planner that selects actions based on current observations and immediate feedback from the environment. 
    \item \textit{Reflexion}~\cite{shinn2023-reflexion}: An extension of \textit{ReAct} that introduces a reflection module. When repeated failures are encountered, it analyzes the execution history and generates a revised plan to guide subsequent decisions. For fair comparison in an online planning setup, we disable environment resets and trigger reflection after three consecutive failed actions.
    \item \textit{ReAct*} and \textit{Reflexion*}: Prompt-augmented variants of the above, in which the LLM receives additional structured descriptions of the task domain, including object types, action semantics, and goal structures. Detailed prompt templates are provided in Appendix~\ref{baselines appendix}.
    \modify{\item \textit{LLM-MCTS}~\cite{zhao2023large-LLM_MCTS}: LLM augmented closed-domain POMDP planning. Given a known object set, LLM generates object-location probability vectors as belief.} It then performs Monte Carlo Tree Search (MCTS), calling the LLM repeatedly to guide the action selection in simulation procedure of MCTS.
\end{itemize}

As shown in Figure~\ref{total results}, \algname significantly outperforms all baselines, demonstrating its capability for planning under uncertainty.

\begin{wrapfigure}{r}{0.33\textwidth}
    \centering
    \includegraphics[width=0.36\textwidth]{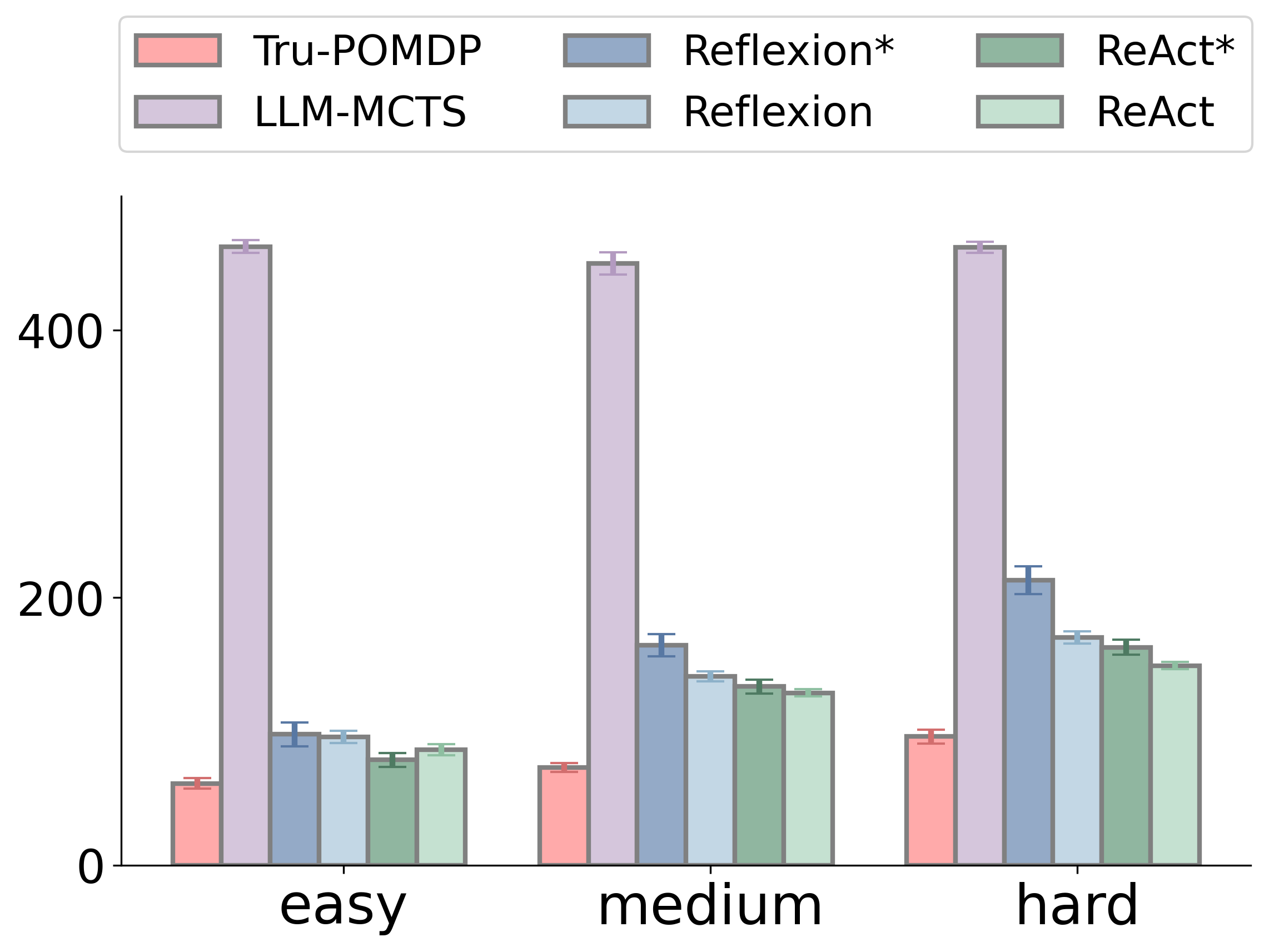}
    \vspace*{-0.5cm}
    \caption{Total tokens (k) used $\downarrow$ by \algname and comparison baselines per episode.}
    \label{exp::tokens}
    \vspace*{-0.5cm}
\end{wrapfigure}

\textbf{\algname vs. LLM-based planners.}
\textit{ReAct} and \textit{Reflexion} are generally ineffective in environments with partial observability. When tasks involve three or more unknown target objects, these methods achieve success rates below 20\% and accumulate negative rewards. The prompt-augmented versions \textit{ReAct*} and \textit{Reflexion*} offer some improvement by providing clearer task semantics to the LLM, increasing success rates to around 40\% and achieving marginally positive rewards. However, their purely reactive nature and lack of principled reasoning over uncertainty limit their performance. In contrast, \algname explicitly reasons over uncertain goals and occluded states using belief-space planning, leading to consistently higher success rates and cumulative rewards.

\textbf{\algname vs. Tree Search planners.}
\textit{LLM-MCTS} struggles with planning efficiency due to frequent LLM calls during simulation, exceeding the 10-minute time limit in all tasks. \algname mitigates this using offline LLM-generated policy, expressed as C++ code.
\modify{\textit{LLM-MCTS} also operates on a closed object set, and can never accomplish a task when the true targets fall outside of the set.
\algname addresses this by constructing open-ended beliefs and dynamic action spaces.}
\algname also integrates LLM reasoning with rigorous belief-space planning and performs tree search over a structured particle belief, enabling it to solve ambiguous tasks more effectively than all baselines. Note that \algname achieved this using minimum consumption of LLM tokens (Figure~\ref{exp::tokens}).
See Appendix~\ref{experiment appendix} for visualization of the planned results.
\subsection{Ablation Study}
\label{ablation_study}

To understand the contributions of key components in \algname, we conduct an ablation study using the following variants:
\begin{itemize}
    \item \textit{w/o Belief}: Removes explicit belief modeling, using only the single most-likely hypothesis generated by the LLM. This variant effectively reduces to an MCTS planner.
    \item \textit{w/o HBU}: Removes the Hybrid Belief Update (HBU), generating an entirely new particle belief from TOH at every step, without Bayesian filtering.
     \item \textit{w/o TOH}: Eliminates the Tree of Hypotheses (TOH) structure, instead querying the LLM once to directly generate a flat set of particle hypotheses, with each particle including a target object set, a goal placement, and an initial placement.
    \item \textit{w/o LRP}: Removes the LLM-Generated Rollout Policy(LRP) in the Belief Tree Search, instead uses a simple policy that randomly selects a legal action.
    \item \textit{w/o BTS}: Removes explicit Belief Tree Search (BTS), directly inputting the particle belief to the LLM and asking it to output the next action.
\end{itemize}

\begin{figure}[t!]
    \centering
    \includegraphics[width=0.9\textwidth]{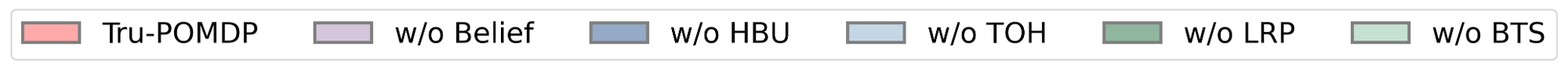}\\[0.5em]
    {\footnotesize  
    \begin{tabular}{@{\hskip 2pt}c@{\hskip 2pt}c@{\hskip 2pt}c@{\hskip 2pt}}
        \includegraphics[width=0.33\textwidth]{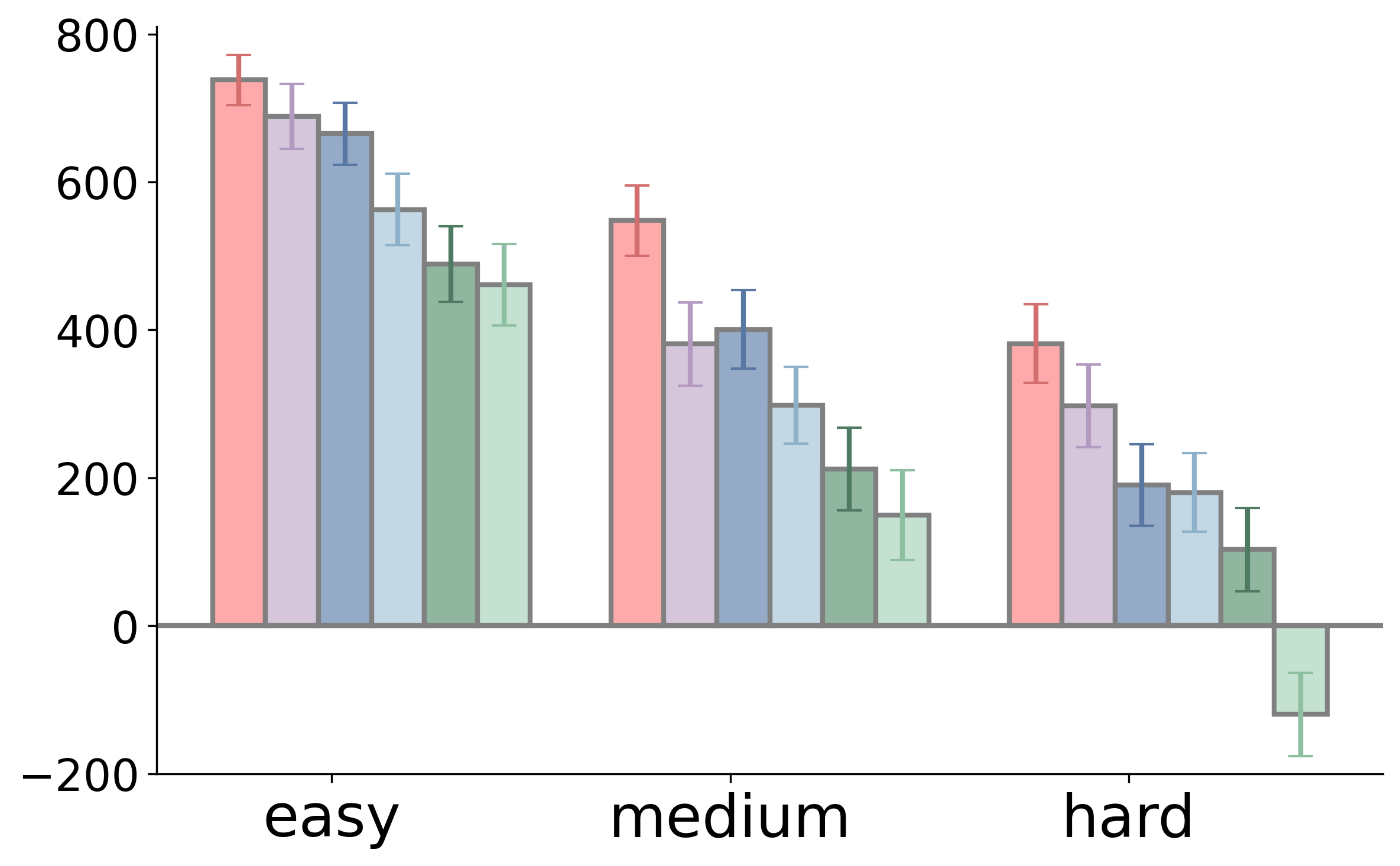} &
        \includegraphics[width=0.33\textwidth]{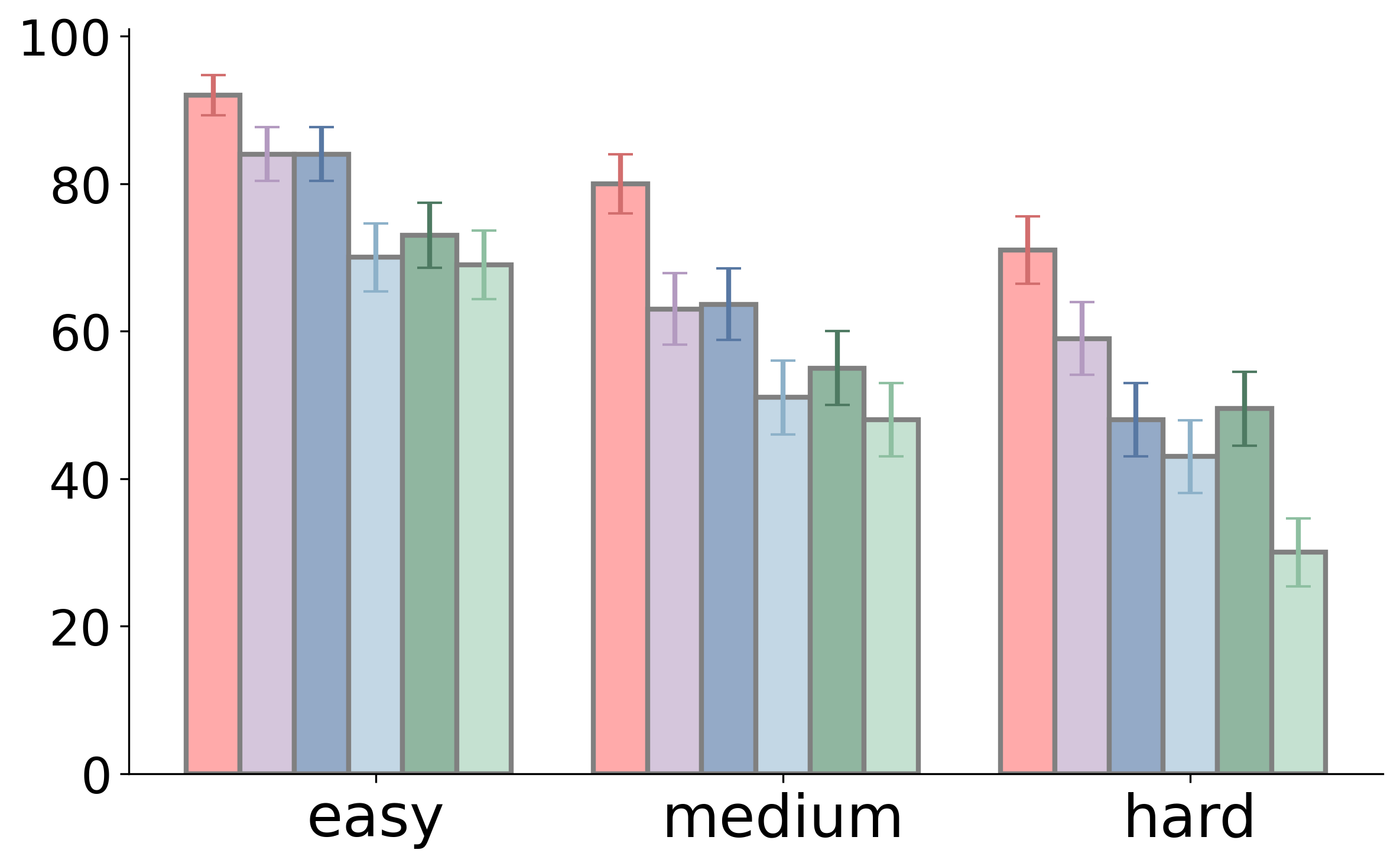} &
        \includegraphics[width=0.33\textwidth]{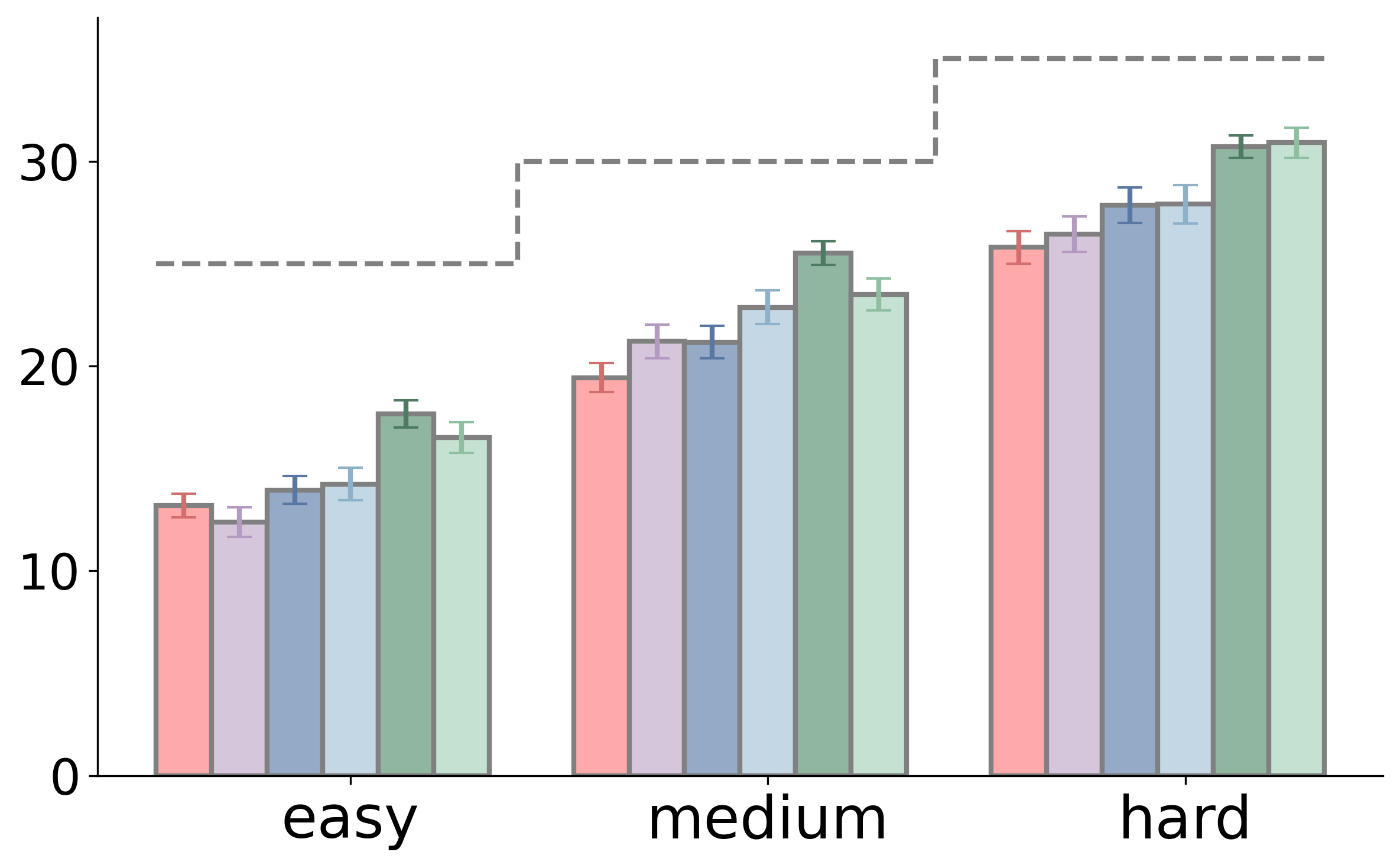} \\
        (a) Cumulative Reward $\uparrow$ &
        (b) Success Rate (\%) $\uparrow$ &
        (c) Total Step Number $\downarrow$
    \end{tabular}
    }
     \caption{Results for Ablation Study. Each bar shows average values with standard error (SE).}
    \label{ablation study}
        \vspace*{-0.5cm}

\end{figure}

As shown in Figure~\ref{ablation study}, the full \algname method consistently outperforms all ablated variants, demonstrating that each proposed module is critical to the overall performance under uncertainty.

Comparing \algname to \textit{w/o Belief}, we observe that explicit belief modeling is essential. Relying solely on a maximum-likelihood hypothesis leads to substantial performance degradation. 
In these scenarios, the level of uncertainty is significant but manageable by proper belief tracking. Without explicit belief representation, the planner fails to recover from errors when the hypotheses diverge from reality.

\begin{wrapfigure}{0}{0.33\textwidth}  
    \centering
    \includegraphics[width=0.36\textwidth]{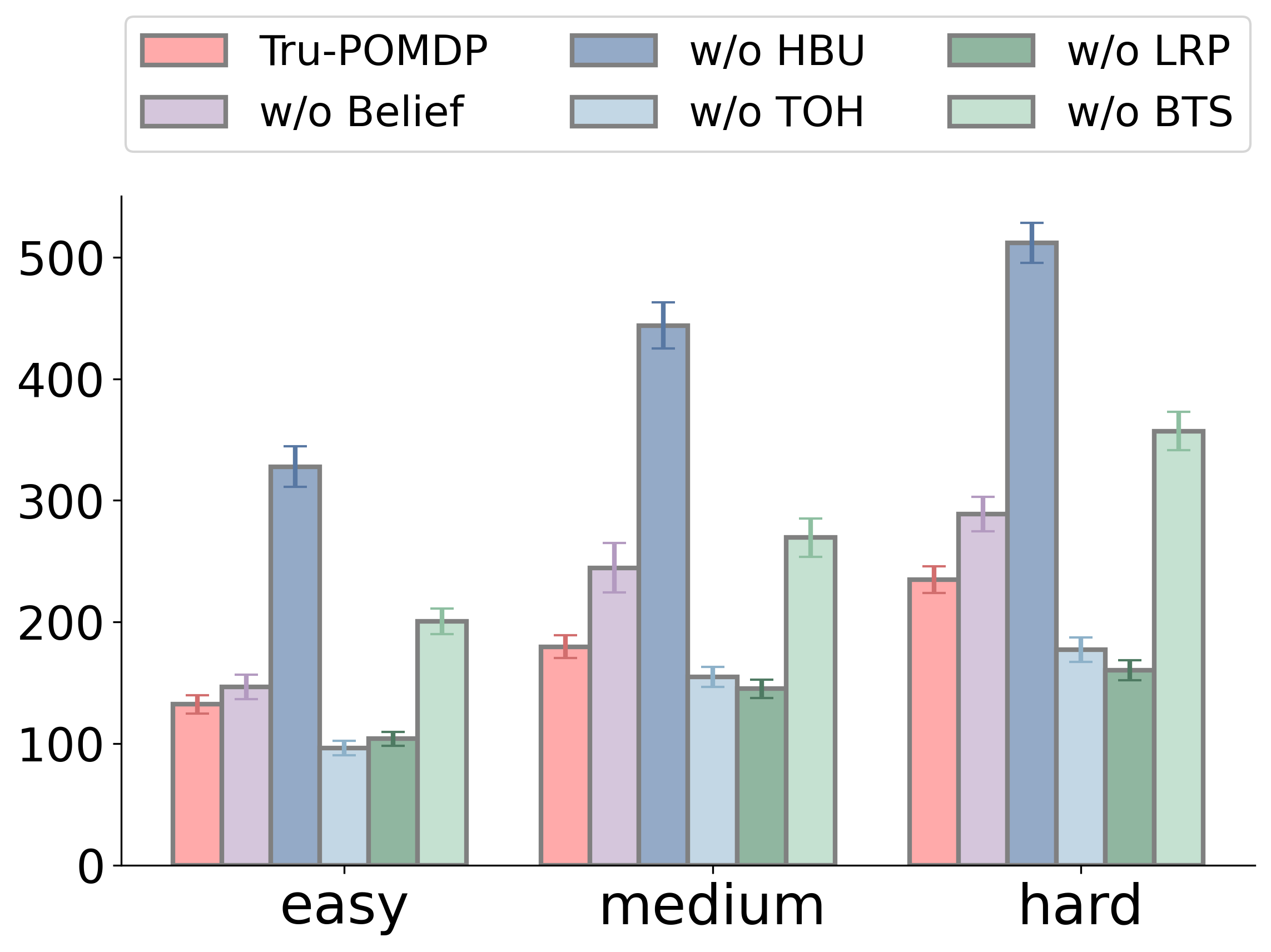}
    \vspace*{-0.5cm}
    \caption{Total planning time $\downarrow$ used by \algname and its ablated variants per episode.}
    \label{ablation time}
    \vspace*{-0.5cm}
\end{wrapfigure}

The comparison against \textit{w/o HBU} highlights the benefit of combining Bayesian filtering with LLM belief generation. 
\draft{Without Bayesian filtering, the performance significantly decreases for tasks with high difficulty where stable belief updates are critical.}
Moreover, Figure~\ref{ablation time} shows that removing Bayesian filtering dramatically increases total planning time, up to three times higher, due to frequent, expensive LLM belief regeneration at every step. In contrast, \algname efficiently maintains and updates beliefs by reusing reliable information from previous steps.

\draft{Comparing \algname with \textit{w/o TOH} highlights the effectiveness of our hierarchical TOH structure. Asking the LLM to generate a flat set of particle hypotheses in a single run degrades belief quality due to hallucination, resulting in worse performance than even the single-hypothesis variant. This underscores the importance of structured querying for reliable belief generation.}

\draft{Comparing \algname with \textit{w/o LRP} shows that the LLM-generated rollout policy offers effective heuristic guidance for tree search. Replacing it with a random policy greatly degrades performance.}

\draft{Comparing with \textit{w/o BTS} confirms the importance of belief-space planning. Removing explicit POMDP planning leads to the lowest rewards and success rates, and the second-longest planning time. This suggests that Tru-POMDP benefits from integration with principled POMDP planning.}

\section{Conclusion and Limitations}
\label{conclusion_and_limitations}

In this work, we introduced \algname, a principled approach that integrates belief-space POMDP planning with structured LLM reasoning to address uncertainties from ambiguous human instructions, open-class objects, and hidden placements. \algname generates explicit particle beliefs via a hierarchical Tree of Hypotheses, refines them through hybrid Bayesian-LLM updates, and plans efficiently with online belief-tree search. Experiments show that this integration significantly outperforms pure LLM and LLM-enhanced tree-search planners on complex household rearrangement tasks, confirming the benefits of explicitly modeling uncertainty and combining POMDP planning with LLM reasoning.

Nevertheless, our approach currently has several limitations. The Tree of Hypotheses incurs computational overhead due to multiple LLM calls, which could be mitigated by fine-tuning an LLM for faster belief generation. 
\modify{Furthermore, \algname leverages conditional independence between the placement of different objects to achieve scalable belief modeling, as this property naively supports factorization. When more complex dependency exists, systematic factorization is required to determine the belief structure.}
Next, our experiments also assumed noise-free perception and deterministic actions. However, the general POMDP formulation naturally accommodates perception noise and stochastic outcomes by adding appropriate probabilistic models. Finally, our evaluation focused on object rearrangement tasks. Extending the approach to larger action spaces and reasoning over additional unknown object attributes remains an important direction for future work.

\section*{Acknowledgment}

This work was supported in part by the National Key R\&D Program of China (Grant No. 2024YFB4707600) and the National Natural Science Foundation of China (Grant No. 62303304).

We used generative AI to improve self-written texts to enhance readability. None of the presented methods and results (figures, equations, numbers, etc.) are generated by AI.

{
\small
\bibliographystyle{unsrt}
\bibliography{references}
}

\newpage

\newpage

\appendix

\section{Technical Appendices}
\label{appendix}
\subsection{Prompt Design for the Tree of Hypotheses}
\label{tree of hypothesis prompt}

\paragraph{Textual Description of the Observation.}
The textual description \(T_z\) of the observation includes: (1) a description of the scene graph \(\mathcal{G}_z\) ; (2) objects that have already reached their respective target areas; and (3) goal states that have been attempted but failed to satisfy the task requirements. Below, we provide an example of such a textual observation.

\begin{lstlisting}
Current Observation: 
The closed areas are: Dishware_Cabinet, Cutlery_Drawer, Cleaning_Supply_Cabinet, Cookware_Cabinet, Bakeware_Cabinet, DishWasher_Inner_Space, Fridge_Cooler_Layer, Fridge_Freezer_Layer, Microwave_Inner_Space, Oven_Inner_Space, 
The open areas are: Human_Hand, Prep_Surface, Coffee_Tea_Surface, Snack_Pantry_Cabinet, Appliance_Surface, Cookbook_Shelf, Pantry_Shelf, Spice_Shelf, Appliance_Cabinet, Drop_Zone_Surface, Beverage_Storage_Cabinet, Display_Surface, Utility_Cabinet, 
The observed objects and their initial areas are: decorative_vase is in Display_Surface, egg_timer is in Coffee_Tea_Surface, gluten_free_cereal is in Human_Hand, granola_bar is in Pantry_Shelf, oatmeal is in Human_Hand, regular_cereal is in Snack_Pantry_Cabinet, rice_jar is in Pantry_Shelf, rolling_pin is in Display_Surface, wheat_bread is in Snack_Pantry_Cabinet, wine_glass is in Beverage_Storage_Cabinet, wine_opener is in Beverage_Storage_Cabinet, 

The wrong goal states are: 
1. gluten_free_cereal in Human_Hand.
2. regular_cereal in Snack_Pantry_Cabinet, gluten_free_cereal in Human_Hand.
3. oatmeal in Human_Hand.
4. gluten_free_cereal in Human_Hand, oatmeal in Human_Hand.

Objects already in target areas: gluten_free_cereal, oatmeal, 
\end{lstlisting}

\paragraph{System Prompt for Level 1 \& 2.}
The prompt consists of five main parts: role, task, guidelines, requirements, and output example. In the requirements section, we design a reasoning chain for the LLM comprising five stages: object of interest identification, target area identification, validation loop, combination generation, and final output certification. The object of interest and target area identification stages extract all target-related objects and their corresponding destination areas. The validation loop checks whether any target-related objects are already in their target areas and can thus be excluded from further consideration. The combination generation stage pairs target objects with target areas to form multiple plausible goal states. Finally, the output certification verifies the legality of the generated results.

\begin{lstlisting}
# Role
You are an assistant to solve an object rearrangement task in a household kitchen environment.
You'll receive:
   1. the language instruction of the task, including the objects_of_interest and their target_areas.
   2. the description of current observation of the environment, listing all open areas, closed areas and observed objects with their placements. If the object is inside a closed area, it is not visible.
   3. A list of previously attempted(wrong) goal states, each containing combinations of objects and target areas that failed to achieve the goal.
   4. a list of objects that have been already placed in the target areas.
# Task
You need to:
1. Identify the correct objects of interest based strictly on the instruction and observation. Only select objects can help complete the task.
2. Select valid target areas for those objects based on the task instruction, using areas explicitly mentioned in the observation.
3. Provide up to k possible valid combinations of objects and their target areas.
# Guidelines
## For objects of interest:
1. The instruction is ambiguous. You need to infer the intent of the language instruction and use common sense.
2. Consider both visual objects and, more importantly, unseen objects of interest in the closed areas.
3. Use '_' to connect multi-word object names (e.g., 'bell_pepper', not 'bell pepper').
4. For the observed objects, use its full name appeared the observation.
## For target areas:
1. The instruction is ambiguous. You need to infer the intent of the language instruction and use common sense.
2. The target areas can only be selected from the areas explicitly mentioned in the observation.
## For objects already in target areas:
1. These objects are checked by human that have been already placed in the target areas.
2. You should totally ignore these objects!!!!!
## For the final answer:
1. You must give out your reasoning process first. Then, you must give your final answer in json format same as the example json answer.
# Critical Rules (Must Read First)
1. Ignore objects already in target areas
   - (Elaborated in Section 3 -> Step1. If an object's current area in the observation equals the designated target area, discard it from consideration.)
2. Cycle Control
   - When returning to Section 1 for re-processing, do not reconsider objects that have already been blacklisted or discarded.
# Requirements: Your reasoning process should include the following sections (Section 1 to 5) and steps in each section. For each step, you should explicitly give out your reasoning and the phased results, and give out your final JSON answer at last.
## Section 1: \objnodeS OF INTEREST IDENTIFICATION
### Step1: \objnodeS OF INTEREST IDENTIFICATION
- Generate up to 10 objects of interest.
- Focus on "fresh" objects not in the blacklist.
- Pay less attention on objects in wrong goal states.
   - The more time the object appeared in wrong goal states, the less attention you should pay to it, and the more probability you should add it to the blacklist.
   - Example: wrong goal states are: 1. chamomile_tea in Spice_Storage_Drawer. Then, pay less attention to chamomile_tea, and consider adding it to the blacklist.
- Totally ignore the objects that have been already placed in the target areas.
   - Example: Objects already in target areas: chamomile_tea, then you should totally ignore chamomile_tea, and move it to the blacklist.
- Object Deficit Resolution Protocol:
  - If the observed objects are insufficient (<= instruction requirements),
    -> Generate hypothetical (unseen) objects that:
      1. Fit the instruction's context and patterns.
      2. Pass semantic coherence checks.
      3. Comply with resource constraints.
      4. Do not duplicate blacklisted properties.
      5. Fulfill missing capabilities in the current object pool.
## Section 2: TARGET \area IDENTIFICATION
### Step1: CONTEXTUAL TARGETING
- Select the most probable target_area for each object.
- Target_area Identification Protocol:
   - Fit the instruction's context and patterns.
   - Reject common-sense conflicts (e.g., placing trash in the refrigerator).
   - Consider Human_Hand first if the instruction is asked for easy access objects or similar intentions.
       - Example intentions: prepare for use, easy to reach, etc.
## Section 3: VALIDATION LOOP
### Step1: TARGET_\area CHECK
- For each candidate object visible in the observation:
  - If object in list that have been already placed in the target areas -> Discard this object entirely.
  - If current area in observation == target_area -> Discard this object entirely.
  - Otherwise, keep it in working memory.
### Step2: COMPLETENESS TEST
- If the remaining objects after discarding cannot fulfill the instruction:
  - Add current candidates to the blacklist
  - Return to Section 1 but exclude blacklisted objects in the next iteration.
## Section 4: COMBINATION GENERATION
### Step1: \objnode-CENTRIC COMBINATION ENGINE
- Generate up to k object-only combinations from the pool of valid objects.
- No target_area assignments yet.
- Each combination must contain no more than 4 objects.
- Prioritize logical groupings and auto-prune duplicates or redundant patterns.
### Step2: POST-HOC TARGET_\area ASSIGNMENT
- For each combination from Step1:
  1. Per-object resolution
     - Select the highest-validity target_area option (per Section 2).
  2. Cross-combination locking
     - The first assignment chosen for an object -> target_area locks that mapping.
     - Subsequent combinations must reuse the same mapping.
### Step3: CROSS-MATRIX VALIDATION
- Consistency Audit
  - Check that every object consistently uses the same target_area in all generated combinations.
- Failure Modes
  - If any target_area mismatch is detected, remove all conflicting combinations.
  - If an object conflict arises, revisit Section 4 step 1 with penalty weighting.
#### Final Safeguards (Section 4)
1. Sequential Locking Protocol
   - The first valid combination's object->target_area assignments bind subsequent combinations.
2. Retroactive Consistency
   - Any new combinations must respect existing locked mappings.
3. Combination Quarantine
   - Combinations involving any unvalidated object-target pair are kept aside until validated.
#### Example (Section 4)
Expected combinations:
  - combination 1: object: blender, target_area: Coffee_Station_Surface; object: cheese_grater, target_area: Flex_Workspace_Surface
  - combination 2: object: blender, target_area: Coffee_Station_Surface; object: potato_peeler, target_area: Daily_Dish_Shelf
Explanation: the same object in 2 combinations (blender) has the same target_area (Coffee_Station_Surface). The combination of objects in 2 combinations are different (blender and cheese_grater, blender and potato_peeler)
Unexpected combinations:
  - combination 1: object: blender, target_area: Coffee_Station_Surface; object: cheese_grater, target_area: Flex_Workspace_Surface
  - combination 2: object: blender, target_area: Daily_Dish_Shelf; object: cheese_grater, target_area: Flex_Workspace_Surface
Explanation: the same object in 2 combinations (blender) has the different target_area (Coffee_Station_Surface and Daily_Dish_Shelf). The combination of objects in 2 combinations are the same (blender and cheese_grater)
## Section 5: FINAL ANSWER
### Step1: RESULT AGGREGATION & VALIDATION
- Combine all validated combinations.
- Explicitly present the final set of object -> target_area mappings.
- Ensure 100% target-area consistency with the locked pairs.
### Step2: FINAL OUTPUT CERTIFICATION
- Only execute after successful validation of sections 1-4.
- Output the final JSON answer if:
  1. All rules in sections 1-4 are satisfied.
  2. Resource allocations remain within bounds.
- The final JSON answer should be the same format with Example output JSON data:
    - Put your json data between ```json and ```
    - Example output JSON data:
```json
{
    "answer": [
        {
            "objects": [
                {
                    "object": "apple",
                    "target_area": "Human_Hand"
                },
                {
                    "object": "banana",
                    "target_area": "robot"
                }
            ],
            "probability": 0.7
        },
        {
            "objects": [
                {
                    "object": "orange",
                    "target_area": "Human_Hand"
                },
                {
                    "object": "banana",
                    "target_area": "robot"
                }
            ],
            "probability": 0.3
        }
    ]
}

\end{lstlisting}

\paragraph{System Prompt for Level 3.}
The prompt adopts the same structure as that used for Levels 1 and 2. The reasoning chain comprises four steps: object visibility check, initial area prediction, consistency verification, and final answer. The object visibility check determines whether the object is observable in the current scene graph \(\mathcal{G}_z\). The initial area prediction and consistency verification steps infer likely initial areas for the target object and ensure that all hypothesized locations for invisible objects are closed areas. The final answer formats the result into the required JSON output schema.

\begin{lstlisting}
# Role
You are an expert assistant specialized in object relocation within household kitchens.
You'll receive:
   1. the language instruction of the task.
   2. the description of current observation of the environment, listing all open areas, closed areas and observed objects with their placements. If the object is inside a closed area, it is not visible.
   3. the name of the object of interest you should now focus on.
# Task
You need to identify up to k most probable initial_areas for every missing object and their probability.
# Guidelines
## For the Initial_areas
1. The object's placement is consistent with common sense.
2. The initial_areas can only be selected from the areas explicitly mentioned in the observation.
## For the final answer:
1. You must give out your reasoning process first. Then, you must give your final answer in json format same as the example json answer.
Now, carefully read the following requirements, then step by step give your reasoning, and finally, generate your answer in JSON format.
# Requirements: Your reasoning process should include the following steps. For each step, you should explicitly give out your reasoning and the phased results.
## Step 1: Object Visibility Check
- Check whether the current object of interest is visible in the observation.
- If the object is visible, set the probability of the object placed in the area to 1.0, and jump to step 4 and give out the final answer.
- If the object's current area is robot's hand, the selected area should be 'robot'.
## Step 2: Object initial_area Guess
- List all the closed areas in the observation.
- Reason/Guess the up to k possible initial_areas for the current object of interest from the closed areas in the observation and corresponding probability using common sense.
## Step 3: Object Initial_Area Double Check
- Double check the initial_areas you proposed:
    1. closed areas in the observation.
    2. The probability sum for all candidate areas must sum to 1.0 for each object.
- Return to step 2 if the double check fails.
## Step 4: Final Answer
- Give out your final JSON answer in the same format of Example Json Answer.
- Example Json Answer:
```json
{
    "answer": [
        {
            "initial_area": "storage_and_inventory_station_surface",
            "probability": 0.7
        },
        {
            "initial_area": "cooler_layer",
            "probability": 0.3
        }
    ]
}
\end{lstlisting}

\subsection{LLM-Generated Rollout Policy in Online POMDP Planning}
\label{rollout policy prompt}

\paragraph{Prompt Design.}
The prompt consists of six main components: role, objective, function signature, available methods, requirements, and final task. The function signature specifies the structure of the policy function, including its name, input parameters, and return type. The available methods section enumerates the callable functions along with detailed usage descriptions, including each method's name, input arguments, return values, and functionality.

\begin{lstlisting}
You are helping a robot perform an object-rearrangement task in a simulated environment. The robot's state is described by a SceneGraphSimple object, which tracks:
1. Which objects and areas exist in the scene.
2. The parent-child relationships between areas and objects.
3. The open/closed status of any given area.
4. Which object, if any, is currently being held by the robot.
There is also a list of goals, where each goal indicates a pairing of [goal_area_id, goal_object_id]. The robot must move each specified object into the specified area.

Your Objective:
Write a C++ function named NextAction that determines the next single action the robot should take to work toward completing any remaining goals. After performing the returned action, the function may be called again until all goals are met.

Function Signature:
ActionSimple NextAction(
    const std::vctor<std::array<int, 2>>& unreached_goals,
    const SceneGraphSimple& current_scene_graph
);
1. unreached_goals is a list of [area_id, object_id] pairs describing which objects still need to be moved to which areas.
2. current_scene_graph describes the state of the environment.

Available Methods in SceneGraphSimple:
You can assume that the following methods exist for querying the scene:
int GetObjectInHand() const;
    // Returns the ID of the object currently in the robot's hand. -1 if none.

bool GetAreaOpenFromId(int area_id) const;
    // Returns true if the specified area is open, false otherwise.

bool CheckAreaInScene(int area_id) const;
    // Returns true if the specified area is known and present in the scene.

bool CheckObjectInScene(int object_id) const;
    // Returns true if the specified object is known and present in the scene.

int GetObjectParent(int object_id) const;
    // Returns the area_id that currently contains this object, or -1 if the object has no parent.

ActionSimple Class and Usage:
enum class ActionType {
    ACTION_OPEN,
    ACTION_PICK,
    ACTION_PLACE
    // ... possibly more types 
};

class ActionSimple {
public:
    ActionSimple();
    ActionSimple(ActionType action_type, int area_id, int object_id);
    // ...
};
1. ActionSimple(ActionType::ACTION_OPEN, area_id, -1) tells the robot to open the specified area.
2. ActionSimple(ActionType::ACTION_PICK, area_id, object_id) instructs the robot to pick up an object from an area.
3. ActionSimple(ActionType::ACTION_PLACE, area_id, -1) instructs the robot to place its currently held object into the specified area.
4. An empty constructor ActionSimple() indicates "no action"-e.g., if there are no remaining tasks.
Requirements:
Return exactly one next action each time the function is called.
The function must compile in a standard C++ environment.
You may use any logic you see fit to decide which action to return next. But pay attentions to:
1. Before taking objects from an closed area, you need to open it first. You can only pick one object at the same time. Generate repeated actions is allowed, but do not generate looped actions.     
2. You need to find the target object and perform the action in the correct order.  You must interact with objects that are observed. You must generate the defined actions.     
If there is no meaningful action to take, you must return ActionSimple() with no arguments (the no-op action).

Final Task:
1. Give out your thinking process, and generate a pseudocode.
2. Check whether your logic is write.
3. Implement the function body for NextAction in valid C++. Do not include extraneous explanations or pseudocode. The function should use the provided parameters and the SceneGraphSimple methods. It should return an ActionSimple that encodes the single, most appropriate action to help achieve the remaining goals.

\end{lstlisting}

\paragraph{LLM-Generated Code for Rollout Policy.}
We send the prompt to the OpenAI \texttt{o1} model and directly adopt the generated code without further modification.

\begin{lstlisting}[language=C++]
ActionSimple NextAction(
    const std::vector<std::array<int, 2>>& unreached_goals,
    const SceneGraphSimple& current_scene_graph
) {
    // If no goals remain, return no-op
    if (unreached_goals.empty()) {
        return ActionSimple();
    }

    // Check each goal
    for (const auto& goal : unreached_goals) {
        int goal_area = goal[0];
        int goal_object = goal[1];

        // Skip if the area or object is not in the scene
        if (!current_scene_graph.CheckAreaInScene(goal_area) ||
            !current_scene_graph.CheckObjectInScene(goal_object)) {
            continue;
        }

        // If goal is already satisfied, skip it
        int parent_of_goal = current_scene_graph.GetObjectParent(goal_object);
        if (parent_of_goal == goal_area) {
            continue;
        }

        // Check what the robot is currently holding
        int object_in_hand = current_scene_graph.GetObjectInHand();

        if (object_in_hand == goal_object) {
            // Need to place it in goal_area
            if (!current_scene_graph.GetAreaOpenFromId(goal_area)) {
                // Open the goal area first
                return ActionSimple(ActionType::ACTION_\open, goal_area, -1);
            } else {
                // Place the object
                return ActionSimple(ActionType::ACTION_\place, goal_area, -1);
            }
        } 
        else if (object_in_hand != -1) {
            // Holding a different object; place it back in its parent
            int parent_of_held = current_scene_graph.GetObjectParent(object_in_hand);
            if (parent_of_held == -1) {
                // No known place to put it; do nothing
                return ActionSimple();
            }
            if (!current_scene_graph.GetAreaOpenFromId(parent_of_held)) {
                return ActionSimple(ActionType::ACTION_\open, parent_of_held, -1);
            } else {
                return ActionSimple(ActionType::ACTION_\place, parent_of_held, -1);
            }
        } 
        else {
            // Holding nothing; pick up the goal object
            if (parent_of_goal == -1) {
                // No parent area known; do nothing
                return ActionSimple();
            }
            if (!current_scene_graph.GetAreaOpenFromId(parent_of_goal)) {
                // Open the parent's area first
                return ActionSimple(ActionType::ACTION_\open, parent_of_goal, -1);
            } else {
                // Then pick the object
                return ActionSimple(ActionType::ACTION_\pick, parent_of_goal, goal_object);
            }
        }
    }

    // If all goals are satisfied or no action can be deduced, return no-op
    return ActionSimple();
}
\end{lstlisting}

\subsection{Kitchen Environments and Area Naming}
\label{kitchen env appendix}

\begin{figure}[h!]
  \centering

  \subfloat[One Wall]{
    \includegraphics[height=3cm]{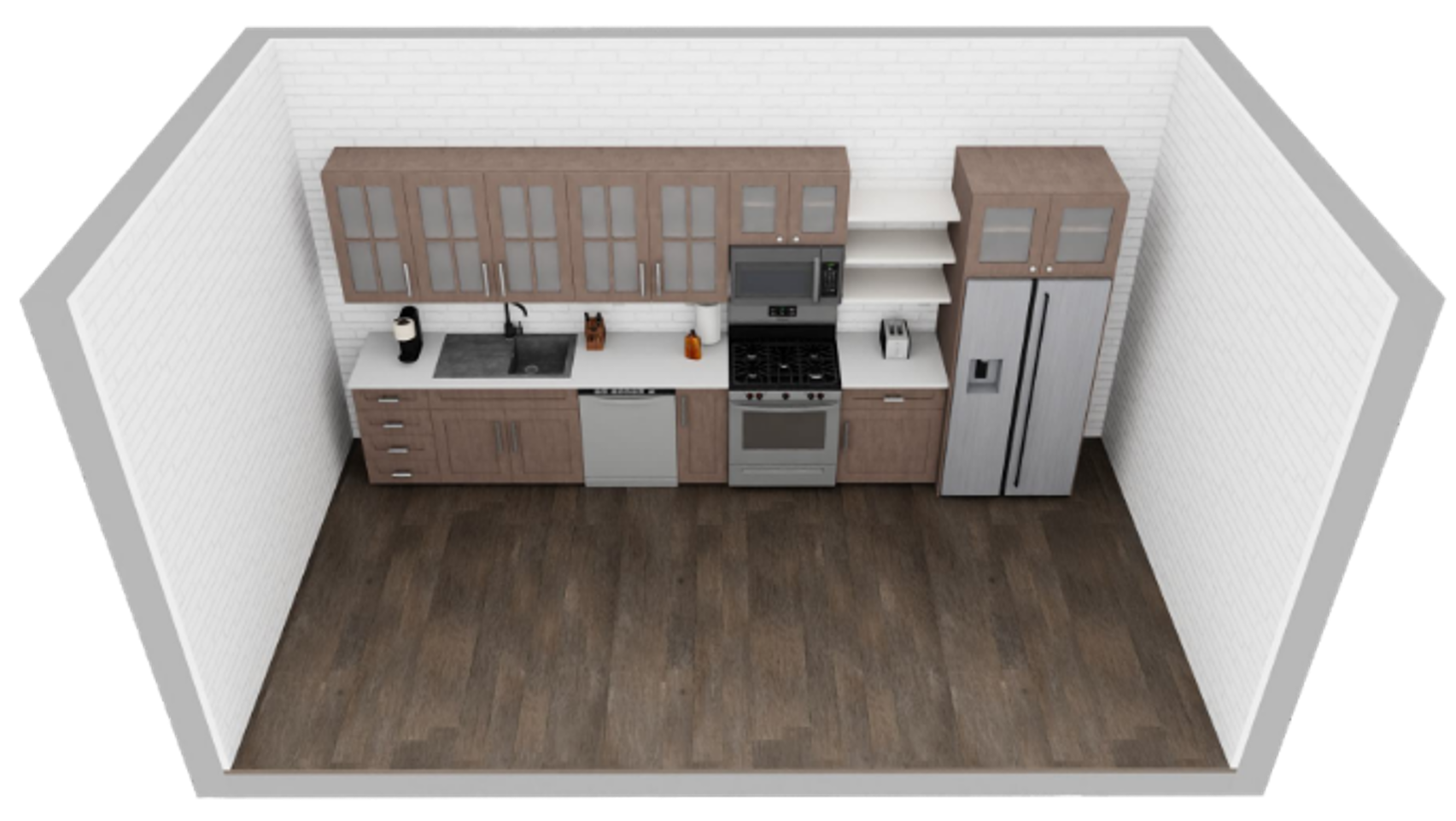}
    \label{one wall}
  }
  \hspace{1em}
  \subfloat[One Wall with Island]{
    \includegraphics[height=3cm]{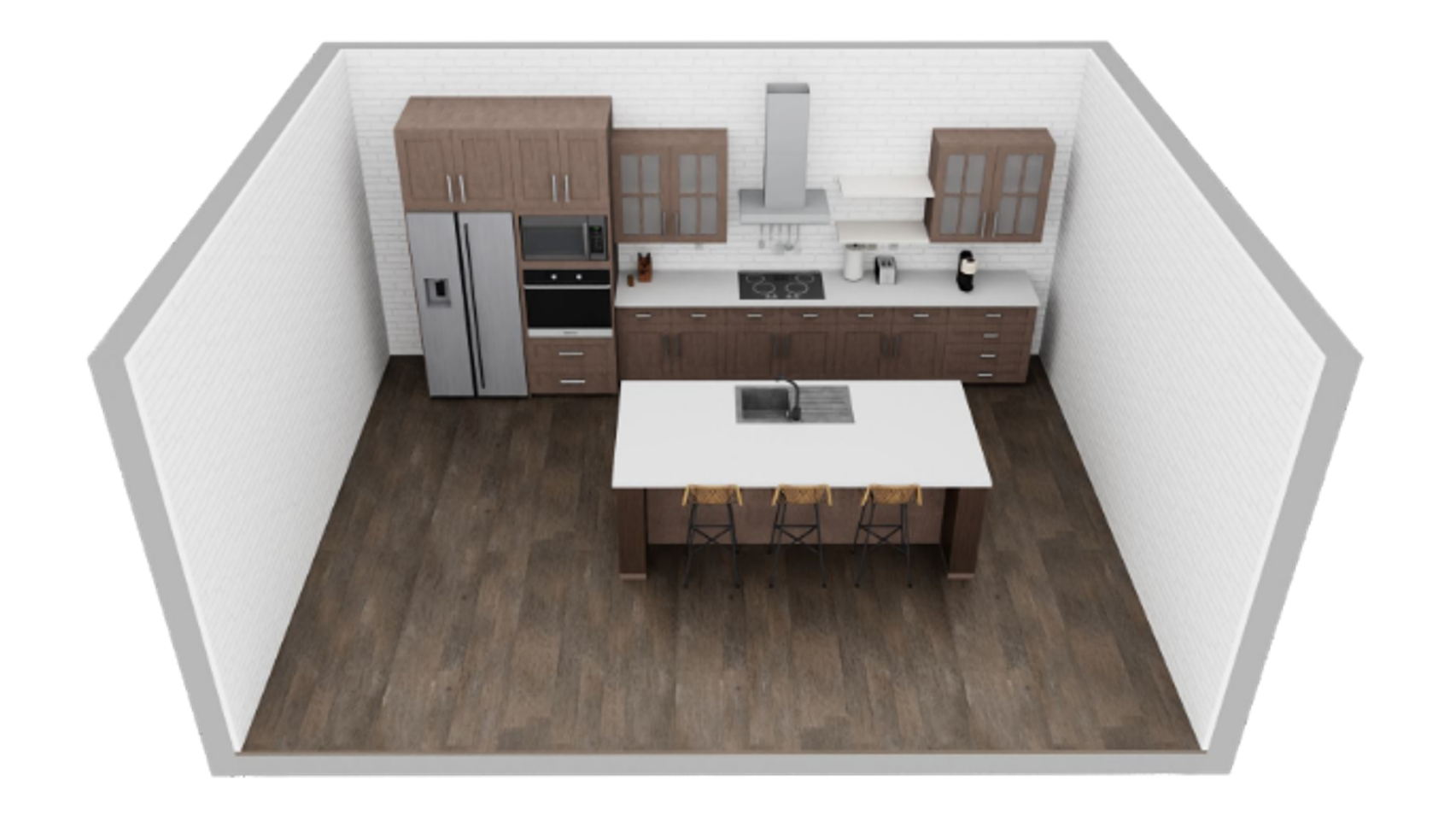}
    \label{one wall large}
  }

  \vspace{1em}

  \subfloat[L-Shaped]{
    \includegraphics[height=3.5cm]{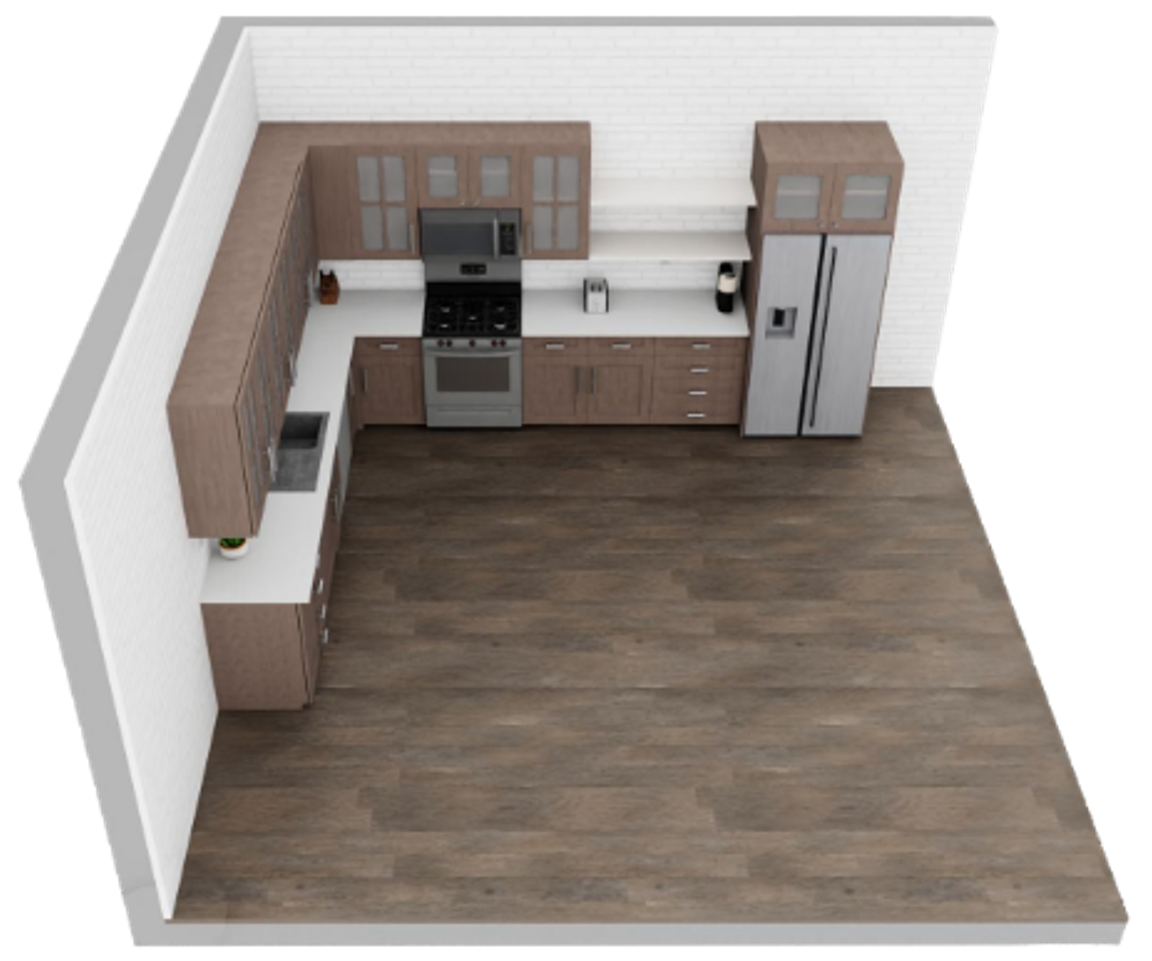}
    \label{l shaped}
  }
  \hspace{1em}
  \subfloat[L-Shaped with Island]{
    \includegraphics[height=3.5cm]{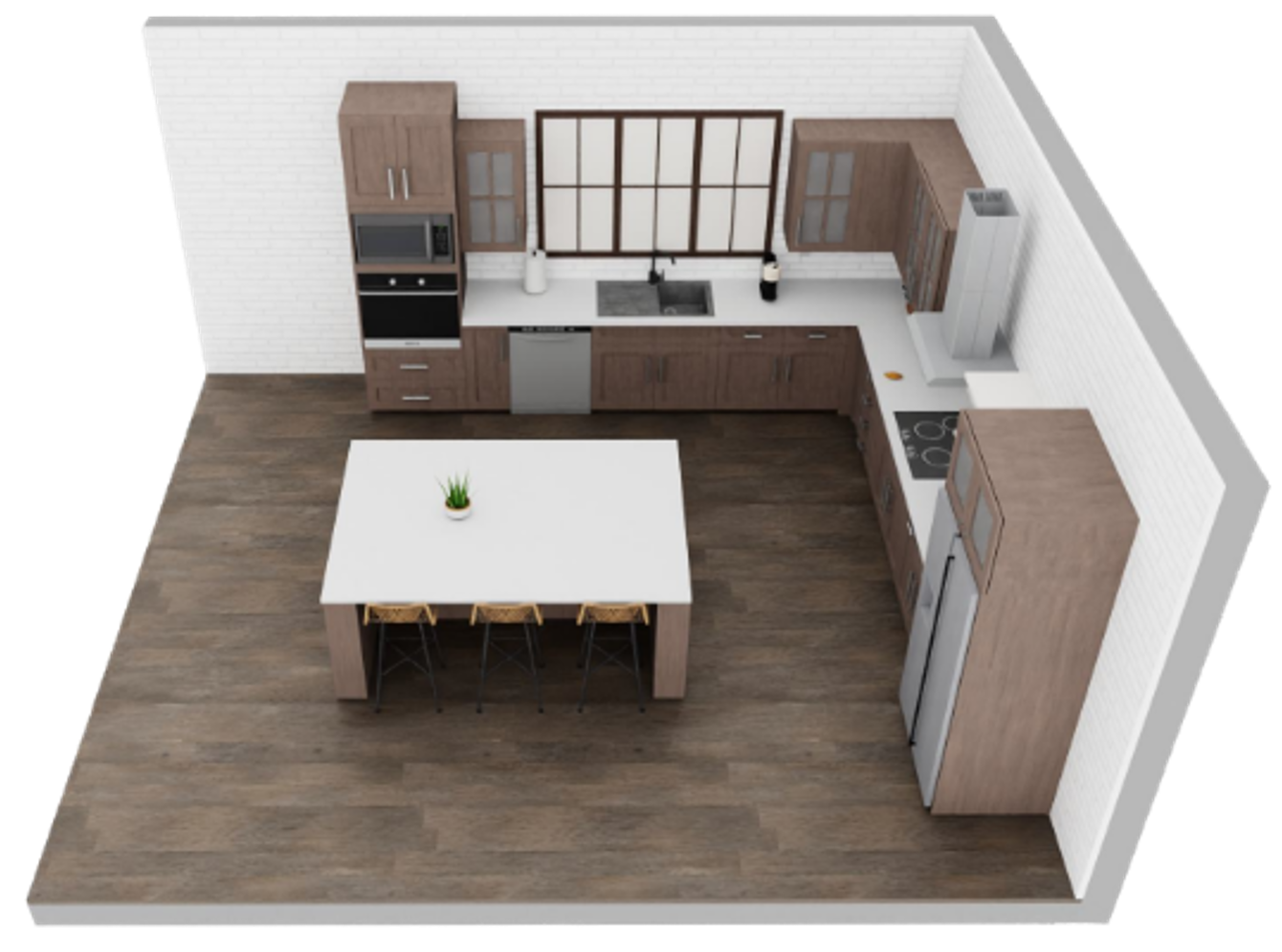}
    \label{l shaped large}
  }
  \hspace{1em}
  \subfloat[Galley]{
    \includegraphics[height=3.5cm]{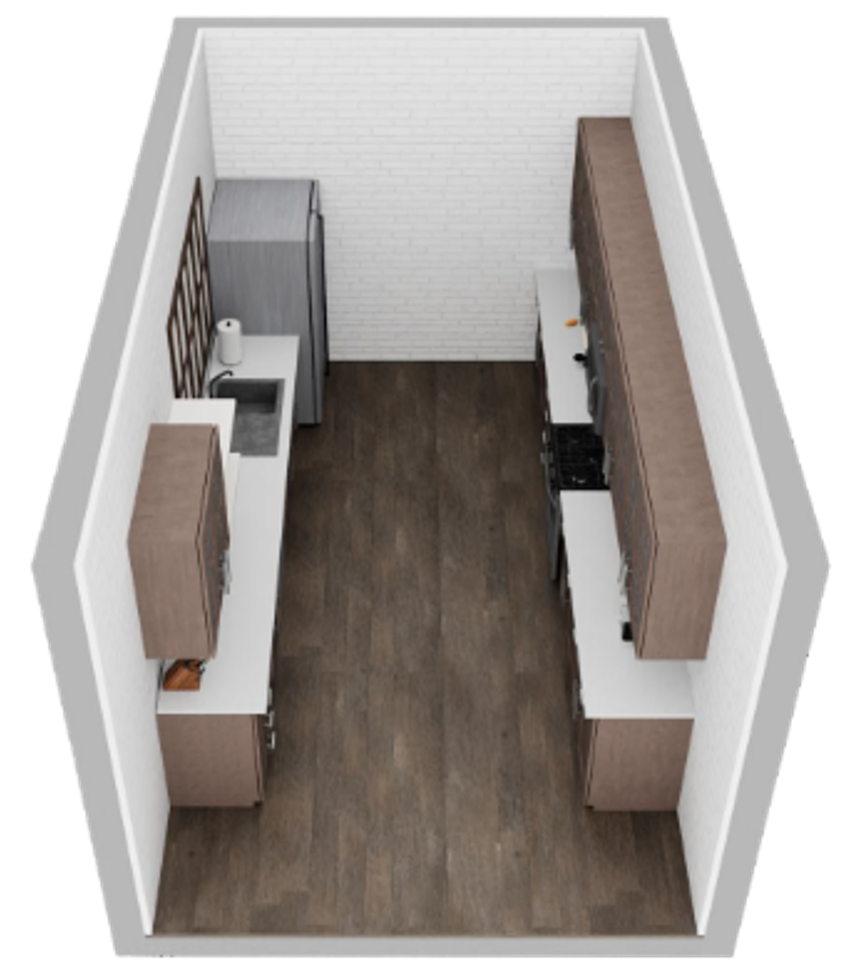}
    \label{galley}
  }

  \caption{Kitchen Environments from RoboCasa~\cite{nasiriany2024robocasa}}.
  \label{fig:kitchen layouts}
\end{figure}

\paragraph{RoboCasa Kitchen Environments.}
We select five kitchen environments from RoboCasa~\cite{nasiriany2024robocasa}. The layouts are illustrated in Figure~\ref{fig:kitchen layouts}. We remove surface-mounted furniture items (e.g., sinks) while retaining furniture with internal storage spaces (e.g., ovens).

\paragraph{Area Naming Scheme.}
We categorize areas in each scene into three types of initially closed areas—furniture internal spaces, cabinets, and drawers—and two types of initially open areas: surfaces and shelves. Below, we list the specific area names for each kitchen environment.

\begin{itemize}
    \item One Wall: 
\begin{table}[H]
\centering
\begin{tblr}{
    width = \textwidth,
    colspec = {Q[1.6cm,c,m] Q[1cm,c,m] X[c,m]}, 
    hlines, vlines,
    cells = {font=\footnotesize}       
}
\textbf{Category} & \textbf{Number} & \textbf{Area Names} \\ 
Furniture Inner spaces & 5 & Fridge\_Cooler\_Layer, Fridge\_Freezer\_Layer, Microwave\_Inner\_Space, Oven\_Inner\_Space, DishWasher\_Inner\_Space \\
Cabinets               & 8 & Dishware\_Cabinet, Snack\_Pantry\_Cabinet, Cleaning\_Supply\_Cabinet, Bakeware\_Cabinet, Cookware\_Cabinet, Utility\_Cabinet, Appliance\_Cabinet, Beverage\_Storage\_Cabinet \\
Drawers                & 1 & Cutlery\_Drawer \\
Surfaces               & 5 & Prep\_Surface, Coffee\_Tea\_Surface, Appliance\_Surface, Display\_Surface, Drop\_Zone\_Surface \\
Shelves                & 3 & Cookbook\_Shelf, Spice\_Shelf, Pantry\_Shelf \\
\end{tblr}
\end{table}

    \item One Wall with Island:
\begin{table}[H]
\centering
\begin{tblr}{
    width = \textwidth,
    colspec = {Q[1.6cm,c,m] Q[1cm,c,m] X[c,m]}, 
    hlines, vlines,
    cells = {font=\footnotesize}       
}
\textbf{Category} & \textbf{Number} & \textbf{Area Names} \\ 
Furniture Inner spaces & 4 & Fridge\_Freezer\_Layer, Fridge\_Cooler\_Layer, Oven\_Inner\_Space, Microwave\_Inner\_Space \\
Cabinets               & 7 & Dry\_Food\_Storage\_Cabinet, Appliance\_Storage\_Cabinet, Baking\_Supplies\_Cabinet, Beverage\_Service\_Cabinet, Cleaning\_Stock\_Cabinet, Bulk\_Storage\_Cabinet, Snack\_Rotation\_Cabinet \\
Drawers                & 8 & Utensil\_Organizer\_Drawer, Knife\_Tool\_Drawer, Spice\_Storage\_Drawer, Wrap\_Container\_Drawer, Towel\_Storage\_Drawer, Utility\_Junk\_Drawer, Trash\_Bag\_Drawer, Quick\_Snack\_Drawer \\
Surfaces               & 5 & Coffee\_Station\_Surface, Prep\_Station\_Surface, Appliance\_Base\_Surface, Breakfast\_Bar\_Surface, Flex\_Workspace\_Surface \\
Shelves                & 2 & Daily\_Dish\_Shelf, Cookbook\_Display\_Shelf \\
\end{tblr}
\end{table}

    \item L-Shaped: 
\begin{table}[H]
\centering
\begin{tblr}{
    width = \textwidth,
    colspec = {Q[1.6cm,c,m] Q[1cm,c,m] X[c,m]}, 
    hlines, vlines,
    cells = {font=\footnotesize}       
}
\textbf{Category} & \textbf{Number} & \textbf{Area Names} \\ 
Furniture Inner spaces & 5 & Fridge\_Freezer\_Layer, Fridge\_Cooler\_Layer, DishWasher\_Inner\_Space, Oven\_Inner\_Space, Microwave\_Inner\_Space \\
Cabinets               & 10 & Cookware\_Storage\_Cabinet, Small\_Appliance\_Cabinet, Backup\_Dishes\_Cabinet, Baking\_Tool\_Cabinet, Cleaning\_Supply\_Cabinet, Dry\_Food\_Pantry\_Cabinet, Beverage\_Storage\_Cabinet, Lunchbox\_Thermos\_Cabinet, Plastic\_Container\_Cabinet, Pet\_Food\_Tool\_Cabinet, Seasonal\_Item\_Cabinet \\
Drawers                & 7 & Cutlery\_Organizer\_Drawer, Cooking\_Tool\_Drawer, Wrap\_Foil\_Drawer, Spice\_Rack\_Drawer, Trash\_Bag\_Recycle\_Drawer, Charging\_Station\_Surface, Recycling\_Sorting\_Surface \\
Surfaces               & 6 & Coffee\_Station\_Surface, Meal\_Prep\_Surface, Small\_Appliance\_Station\_Surface, Breakfast\_Zone\_Surface, Herb\_Garden\_Surface, Flex\_Workspace\_Surface \\
Shelves                & 2 & Daily\_Spice\_Shelf, Cookbook\_Display\_Shelf \\
\end{tblr}
\end{table} 

    \item L-Shaped with Island: 
\begin{table}[H]
\centering
\begin{tblr}{
    width = \textwidth,
    colspec = {Q[1.6cm,c,m] Q[1cm,c,m] X[c,m]}, 
    hlines, vlines,
    cells = {font=\footnotesize}       
}
\textbf{Category} & \textbf{Number} & \textbf{Area Names} \\ 
Furniture Inner spaces & 5 & Oven\_Inner\_Space, Microwave\_Inner\_Space, Dishwasher\_Inner\_Space, Fridge\_Freezer\_Layer, Fridge\_Cooler\_Layer \\
Cabinets               & 10 & Pantry\_Ingredient\_Cabinet, Dishware\_Cabinet, Cookware\_Cabinet, Cleaning\_Supply\_Cabinet, Baking\_Tool\_Cabinet, Pet\_Food\_Tool\_Cabinet, Waste\_Management\_Cabinet, Appliance\_Storage\_Cabinet, Beverage\_Cabinet, Spice\_Jar\_Cabinet \\
Drawers                & 6 & Utensil\_Drawer, Foodwrap\_Drawer, Condiment\_Packet\_Drawer, Knife\_Block\_Drawer, Baking\_Mold\_Drawer, Tea\_Coffee\_Drawer \\
Surfaces               & 7 & Small\_Appliance\_Surface, Prep\_Surface, Coffee\_Station\_Surface, Daily\_Seasoning\_Surface, Fruit\_Basket\_Surface, Kitchen\_Tool\_Surface, Decorative\_Surface \\
Shelves                & 2 & Cookbook\_Shelf, Display\_Shelf \\
\end{tblr}
\end{table}

    \item Galley:
\begin{table}[H]
\centering
\begin{tblr}{
    width = \textwidth,
    colspec = {Q[1.6cm,c,m] Q[1cm,c,m] X[c,m]}, 
    hlines, vlines,
    cells = {font=\footnotesize}       
}
\textbf{Category} & \textbf{Number} & \textbf{Area Names} \\ 
Furniture Inner spaces & 5 & Fridge\_Freezer\_Layer, Fridge\_Cooler\_Layer, Dishwasher\_Inner\_Space, Oven\_Inner\_Space, Microwave\_Inner\_Space \\
Cabinets               & 12 & Dry\_Food\_Storage\_Cabinet, Small\_Appliance\_Cabinet, Liquor\_Storage\_Cabinet, Cleaning\_Supply\_Cabinet, Backup\_Dishes\_Cabinet, Baking\_Ingredients\_Cabinet, Bulk\_Food\_Cabinet, Seasonal\_Items\_Cabinet, Cookware\_Storage\_Cabinet, Servingware\_Display\_Cabinet, Pet\_Supply\_Cabinet, Medicine\_Storage\_Cabinet \\
Drawers                & 11 & Utensil\_Organizer\_Drawer, Cooking\_Tool\_Drawer, Spice\_Rack\_Drawer, Foil\_Wrap\_Drawer, Snack\_Storage\_Drawer, Knife\_Block\_Drawer, Kitchen\_Linen\_Drawer, Recycling\_Bin\_Drawer, Lunch\_Container\_Drawer, Tea\_Coffee\_Drawer, preserved\_ingredients\_drawer \\
Surfaces               & 10 & Coffee\_Station\_Surface, Chopping\_Station\_Surface, Breakfast\_Prep\_Surface, Rice\_Cooker\_Surface, Mixing\_Station\_Surface, Baking\_Prep\_Surface, Fruit\_Basket\_Surface, Microwave\_Station\_Surface, Knife\_Magnet\_Surface, Herb\_Garden\_Surface \\
Shelves                & 2 & Cookbook\_Display\_Shelf, Daily\_Dishes\_Shelf \\
\end{tblr}
\end{table}

\end{itemize}

\subsection{LLM-Assisted Task Generation}
\label{task generation}

We generate object rearrangement tasks with the assistance of an LLM (GPT-4o). Each task comprises four components: a natural language instruction; target-related objects with their initial placements; disturbance objects with their initial placements; and a set of goal states that fulfill the instruction's requirements.

First, we sample a background context for the task from a predefined set of character and temporal settings. The LLM is then prompted to generate an ambiguous language instruction that a human might provide, along with 6–8 target-related objects and their initial areas. These initial placements are deliberately chosen not to satisfy the instruction's requirements directly, ensuring that rearrangement is necessary.

Next, we query the LLM to generate additional disturbance objects unrelated to the task, increasing the total number of objects in the scene to 20. For each disturbance object, the LLM is asked to propose four plausible initial areas based on commonsense priors, from which we uniformly sample one as its placement.

Finally, we prompt the LLM to generate up to four plausible target areas for each target-related object and to enumerate all valid combinations of target-related objects that satisfy the task requirements. We then compute all potential goal states, each comprising the target objects and their corresponding target areas.

An example JSON file of a generated task is shown below.

\begin{lstlisting}
{
    "placement": [
        {
            "area": "Plastic_Container_Cabinet",
            "placed_object": "gluten_free_flour_15"
        },
        {
            "area": "Seasonal_Item_Cabinet",
            "placed_object": "sugar_cookie_molds_20"
        },
        {
            "area": "Plastic_Container_Cabinet",
            "placed_object": "dinner_plate_15"
        },
        {
            "area": "Lunchbox_Thermos_Cabinet",
            "placed_object": "leftover_container_5"
        },
        {
            "area": "Cookware_Storage_Cabinet",
            "placed_object": "rice_cooker_14"
        },
        {
            "area": "Cooking_Tool_Drawer",
            "placed_object": "serving_spoon_3"
        },
        {
            "area": "Breakfast_Zone_Surface",
            "placed_object": "measuring_cup_4"
        },
        {
            "area": "Small_Appliance_Cabinet",
            "placed_object": "mug_warmer_10"
        },
        {
            "area": "Beverage_Storage_Cabinet",
            "placed_object": "wine_opener_9"
        },
        {
            "area": "Breakfast_Zone_Surface",
            "placed_object": "egg_timer_11"
        },
        {
            "area": "Cookbook_Display_Shelf",
            "placed_object": "scented_candle_3"
        },
        {
            "area": "Beverage_Storage_Cabinet",
            "placed_object": "herb_press_5"
        },
        {
            "area": "Seasonal_Item_Cabinet",
            "placed_object": "candle_3"
        },
        {
            "area": "Meal_Prep_Surface",
            "placed_object": "Fruit_Basket_7"
        },
        {
            "area": "Cookbook_Display_Shelf",
            "placed_object": "decorative_vase_6"
        },
        {
            "area": "Beverage_Storage_Cabinet",
            "placed_object": "wine_glass_13"
        },
        {
            "area": "Backup_Dishes_Cabinet",
            "placed_object": "sushi_rolling_mat_12"
        },
        {
            "area": "Small_Appliance_Station_Surface",
            "placed_object": "decorative_vase_8"
        },
        {
            "area": "Small_Appliance_Cabinet",
            "placed_object": "wine_opener_4"
        },
        {
            "area": "Breakfast_Zone_Surface",
            "placed_object": "coffee_mug_9"
        },
        {
            "area": "Spice_Rack_Drawer",
            "placed_object": "candlestick_holder_6"
        }
    ],
    "robot": {
        "name": "Fetch",
        "location": "Cookbook_Display_Shelf",
        "object_in_hand": ""
    },
    "task": {
        "instruction": "As the evening wind-down begins and focusing on allergen safety, please ensure gluten-free baking supplies are accessible but separate from common utensils while also organizing any dinner leftovers for tomorrow's lunch.",
        "goal_set": [
            [
                {
                    "target_area": "Baking_Tool_Cabinet",
                    "placed_object": "gluten_free_flour_15"
                },
                {
                    "target_area": "Fridge_Cooler_Layer",
                    "placed_object": "leftover_container_5"
                }
            ],
            [
                {
                    "target_area": "Baking_Tool_Cabinet",
                    "placed_object": "gluten_free_flour_15"
                }
            ],
            [
                {
                    "target_area": "Dry_Food_Pantry_Cabinet",
                    "placed_object": "gluten_free_flour_15"
                },
                {
                    "target_area": "Fridge_Cooler_Layer",
                    "placed_object": "leftover_container_5"
                }
            ],
            [
                {
                    "target_area": "Dry_Food_Pantry_Cabinet",
                    "placed_object": "gluten_free_flour_15"
                }
            ]
        ]
    }
}

\end{lstlisting}


\subsection{Hyperparameters in \algname}
\label{hyper parameters}

\begin{table}[H]
\centering
\begin{tblr}{
    width = \textwidth,
    colspec = {Q[1.5cm,c,m] X[c,m] Q[1cm,c,m]},
    hlines, vlines,
    row{1} = {font=\bfseries},
    cells = {font=\small}
}
\textbf{Parameter} & \textbf{Description} & \textbf{Value} \\ 
$C_1, C_2$ & Number of candidates for Levels 1 and 2 in the Tree of Hypotheses & 3 \\
$T$ & Temperature for the LLM in the Tree of Hypotheses & 0.1 \\
$\epsilon$ & Threshold in Hybrid Belief Update & 0.7 \\
$k$ & Number of scenarios in Belief Tree Search & 30 \\
$d_s$ & Maximum search depth in Belief Tree Search & 20 \\
$d_r$ & Rollout policy execution depth & 10 \\
\end{tblr}
\end{table}

\subsection{Baseline Modifications}
\label{baselines appendix}

\paragraph{Online Planning Reflexion.}
Reflexion was originally designed for offline planning. We therefore introduce the following modifications for the online planning setting. First, we revise the activation mechanism of the reflection module: when the LLM either produces a thinking process three times consecutively or generates three infeasible actions in a row, we consider the LLM unable to provide valid outputs. At that point, the reflection module is triggered to generate a new plan for guidance. Additionally, we restrict the history record to the last 10 time steps to reduce input token length and LLM query latency.

\paragraph{System Prompt for ReAct* and Reflexion*.}
In the original implementation, ReAct and Reflexion used GPT-3.5 for simple tasks, guiding the LLM solely via multiple demonstrations. In our evaluation, we observe that this approach performs poorly when switching to the GPT-4.1 and applying to more complex tasks. We attribute this to the increased complexity of observations and stricter action constraints. To address this, we design ReAct* and Reflexion*, which incorporate an additional system prompt specifying the role, task, action constraints, and other relevant details.

\begin{lstlisting}
# Role:
You are an robot to complte the kitchenware household tasks. At first, you'll receive
1. current observation of the environment.
2. the task you need to complete.
At each step, you'll receive
2. current observation of the environment.

# Task:
You need to give the next best action. The legal actions are:
1. Open(area): open the door of an closed area
2. Pick(area, object): pick the observed object from an open area
3. Place(area): place the object in robot's hand in/on an open area

# Note:
1. If an area is open, it's fully observable, don't try to explore the open areas.
2. If you're still been asked to complete the task (receive action feedback), it means you haven't complete the task yet. Please try other solutions.
3. The action must follow the given format, and the area and object parameters in the action must appeared in the observation.
4. You can generate 2 kinds of answers.
    (1) thinking process: If you want to think for the current step, you should begin you answer with "think: ". At this process, you don't need to generate the next best action. 
    (2)next best action: If you received 'ok.' at the last of the interaction history, it means that you have thought about the current step. Therefore, please answer one action directly. 
5. please don't generate '>' or anything else in front of your answer.
\end{lstlisting}

\subsection{Visualization of Planned Results}
\label{experiment appendix}
\begin{figure}[H]
    \centering
    \includegraphics[width=1\linewidth]{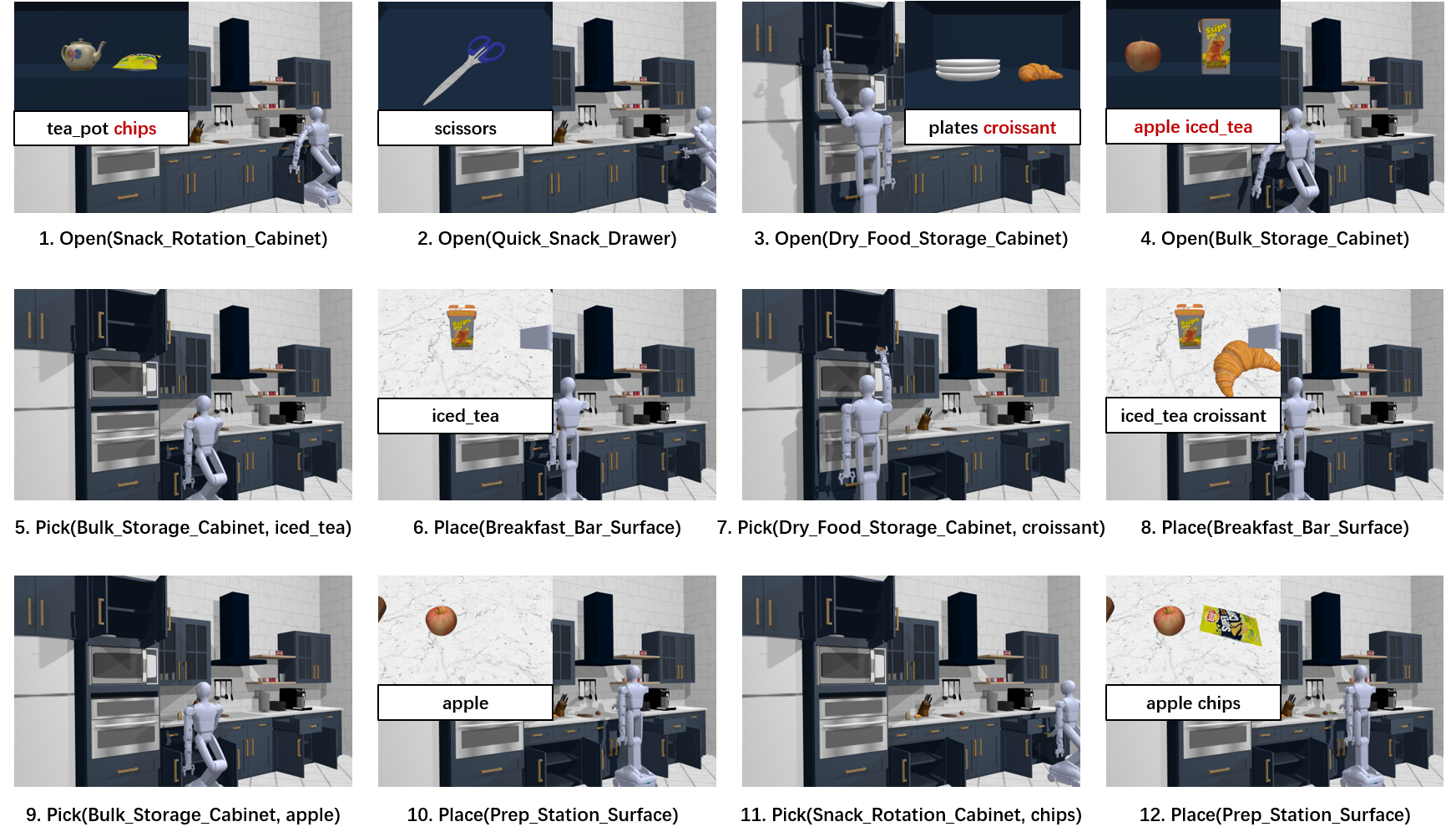}
    \caption{Visualization of planned results.} 
    \label{planned results}
\end{figure}

We visualize an example of the planned results in Figure~\ref{planned results}. In this task, the instruction is: \textit{“I'm preparing for my big housewarming gathering this afternoon. Could you help me sort out items to ensure that snacks are clearly visible?”} The robot identifies all target objects (chips, croissant, apple, and iced\_tea) within 4 steps and completes the rearrangement by placing all target objects onto surfaces within 8 steps.

\end{document}